\documentclass[lettersize,journal]{IEEEtran}

\usepackage{amsmath,amsfonts,bm}

\def\eqref#1{equation~\ref{#1}}

\def\1{\bm{1}}

\def\rva{{\mathbf{a}}}

\def\rvg{{\mathbf{g}}}

\def\rvs{{\mathbf{s}}}

\def\mM{{\bm{M}}}

\DeclareMathAlphabet{\mathsfit}{\encodingdefault}{\sfdefault}{m}{sl}
\SetMathAlphabet{\mathsfit}{bold}{\encodingdefault}{\sfdefault}{bx}{n}
\newcommand{\tens}[1]{\bm{\mathsfit{#1}}}

\def\tG{{\tens{G}}}

\def\gA{{\mathcal{A}}}

\def\gD{{\mathcal{D}}}

\def\gG{{\mathcal{G}}}

\def\gL{{\mathcal{L}}}

\def\gP{{\mathcal{P}}}

\def\gR{{\mathcal{R}}}
\def\gS{{\mathcal{S}}}
\def\gT{{\mathcal{T}}}

\def\sG{{\mathbb{G}}}

\def\sM{{\mathbb{M}}}

\def\sT{{\mathbb{T}}}

\newcommand{\E}{\mathbb{E}}

\newcommand{\R}{\mathbb{R}}

\DeclareMathOperator*{\argmax}{arg\,max}
\DeclareMathOperator*{\argmin}{arg\,min}

\newcommand{\te}[1]{\texttt{#1}}

\usepackage{amsmath,amsfonts}
\usepackage{algorithmic}
\usepackage{algorithm}
\usepackage{array}
\usepackage{textcomp}
\usepackage{stfloats}
\usepackage{url}
\usepackage{verbatim}
\usepackage{graphicx}
\usepackage{cite}
\usepackage{multirow}
\usepackage{colortbl}
\usepackage{xcolor}
\usepackage{float}
\usepackage[pagebackref,breaklinks,colorlinks]{hyperref}
\hypersetup{citecolor=[RGB]{119,185,0}}
\usepackage[capitalize,noabbrev]{cleveref}
\usepackage[numbers, compress]{natbib}
\usepackage{breakcites}
\usepackage{amssymb}
\usepackage[T1]{fontenc}
\usepackage{duckuments}
\usepackage{tabularx,booktabs}
\usepackage{pifont}
\usepackage{wrapfig}
\usepackage{enumitem}
\usepackage{subfigure}
\newtheorem{theorem}{Theorem}[section]

\newtheorem{definition}[theorem]{Definition}

\usepackage{bbding}

\begin{document}

\title{Task-Aware Harmony Multi-Task Decision Transformer for
Offline Reinforcement Learning}

\author{Ziqing Fan*,
~Shengchao~Hu*,
~Yuhang Zhou,
        Li~Shen, 
        Ya Zhang, 
        Yanfeng Wang, ~and~Dacheng~Tao,~\IEEEmembership{Fellow,~IEEE}%
\thanks{Ziqing Fan, Shengchao Hu, Yuhang Zhou, Ya Zhang and Yanfeng Wang are with Shanghai Jiao Tong University and Shanghai AI Lab, China. Email: \{zqfan\_knight, charles-hu, zhouyuhang, ya\_zhang, wangyanfeng\}@sjtu.edu.cn}
\thanks{Li Shen is with Shenzhen Campus of Sun Yat-sen University, Shenzhen 518107, China.  Email: mathshenli@gmail.com}
\thanks{Dacheng Tao is with Nanyang Technological University, Singapore.
Email: dacheng.tao@ntu.edu.sg}
\thanks{Corresponding author: Li Shen and Yanfeng Wang}
}

\maketitle

\begin{abstract}
    The purpose of offline multi-task reinforcement learning (MTRL) is to develop a unified policy applicable to diverse tasks without the need for online environmental interaction.
    Recent advancements approach this through sequence modeling, leveraging the Transformer architecture's scalability and the benefits of parameter sharing to exploit task similarities.
    However, variations in task content and complexity pose significant challenges in policy formulation, necessitating judicious parameter sharing and management of conflicting gradients for optimal policy performance.
    Furthermore, identifying the optimal parameter subspace for each task often necessitates prior knowledge of the task identifier during inference, limiting applicability in real-world scenarios with variable task content and unknown current tasks.
    In this work, we introduce the Harmony Multi-Task Decision Transformer (HarmoDT), a novel solution designed to identify an optimal harmony subspace of parameters for each task. 
    We formulate this as a bi-level optimization problem within a meta-learning framework, where the upper level learns masks to define the harmony subspace, while the inner level focuses on updating parameters to improve the overall performance of the unified policy.
    To eliminate the need for task identifiers, we further design a group-wise variant (G-HarmoDT) that clusters tasks into coherent groups based on gradient information, and utilizes a gating network to determine task identifiers during inference.
    Empirical evaluations across various benchmarks highlight the superiority of our approach, demonstrating its effectiveness in the multi-task context with specific improvements of 8\% gain in task-provided settings, 5\% in task-agnostic settings, and 10\% in unseen settings.
\end{abstract}


\section{Introduction}
\IEEEPARstart{O}{ffline} reinforcement learning (RL) \citep{levine2020offline} enables the learning of policies directly from an existing offline dataset, thus eliminating the need for interaction with the actual environment.
Despite the promising developments of offline RL in various robotic tasks, its successes have been largely confined to individual tasks within specific domains, such as locomotion or manipulation \citep{fu2020d4rl, CQL, CommFormer}.
Drawing inspiration from human learning capabilities, where individuals often acquire new skills by building upon existing ones and spend less time mastering similar tasks, there's a growing interest in the potential of training a set of tasks with inherent similarities in a more cohesive and efficient manner \citep{lee2022multi}.
This perspective leads to the exploration of multi-task reinforcement learning (MTRL), which seeks to develop a versatile policy capable of addressing a diverse range of tasks.

\begin{figure}[!t]
\centering
\vspace{5pt}
\includegraphics[width=0.42\textwidth]{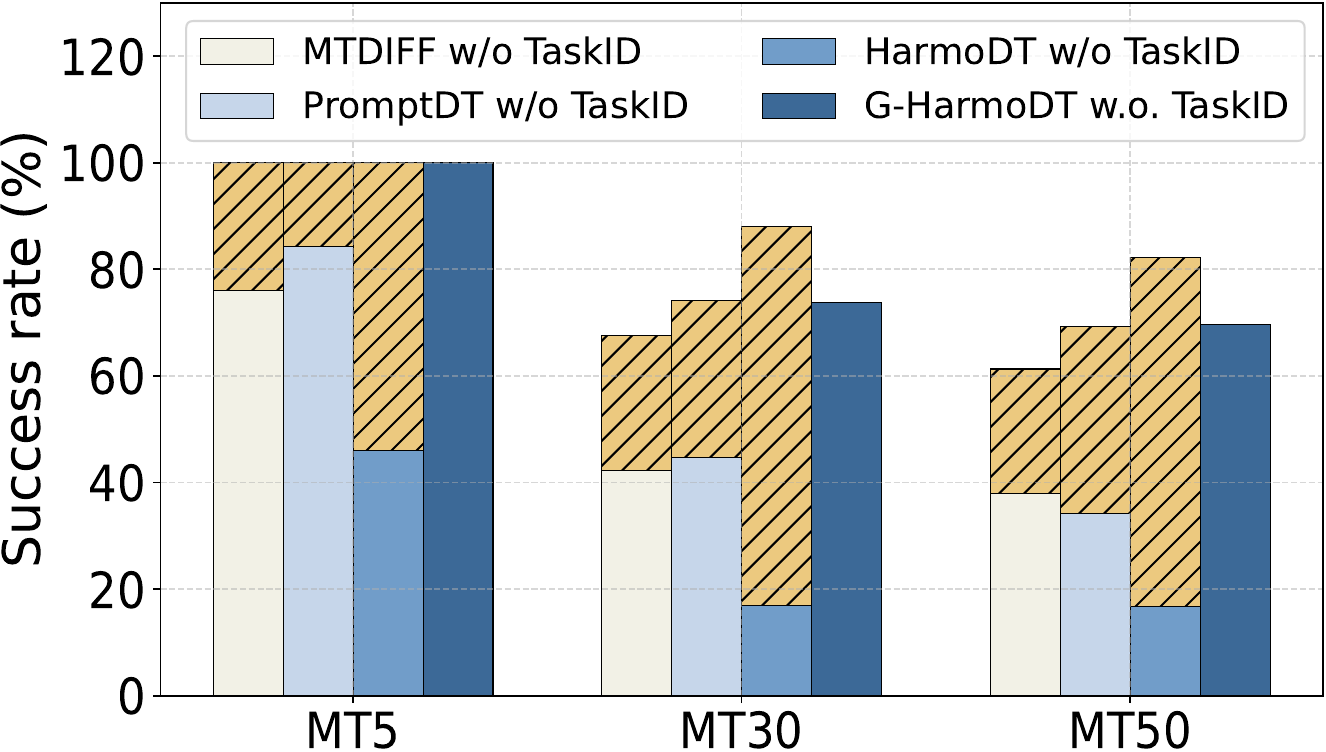}
\caption{
Accuracy drops as the number of tasks increases and task identifiers are absent in near-optimal cases within the Meta-World benchmark, focusing on a comparison of our method with prevalent MTRL algorithms, PromptDT and MTDIFF.
More results of baselines and analysis refer to Section~\ref{sec:exp}.
Methods including HarmoDT face significant performance drops without task identifiers, whereas our G-HarmoDT effectively overcomes this limitation.}
\label{fig:performance_num}
\vspace{-0.2cm}
\end{figure}

Recent developments in Offline RL, such as the Decision Transformer \citep{DT} and Trajectory Transformer \citep{TT}, have abstracted offline RL as a sequence modeling (SM) problem, showcasing their ability to transform extensive datasets into powerful decision-making tools \citep{TRL}. 
These models are particularly beneficial for multi-task RL challenges, offering a high-capacity framework capable of accommodating task variances and assimilating extensive knowledge from diverse datasets. 
Additionally, they open up possibilities for integrating advancements \citep{brown2020language} from language modeling into MTRL methodologies.
However, the direct application of these high-capacity sequential models to MTRL presents considerable algorithmic challenges.
As indicated by \citet{yu2020meta}, simply employing a shared network backbone for all diverse robot manipulation tasks can lead to severe gradient conflicts.
This situation arises when the gradient direction for a particular task starkly contrasts with the majority consensus direction.
Such unregulated sharing of parameters and their optimization under conflicting gradient conditions can contravene the foundational goals of MTRL, degrading performance compared to task-specific training methods \citep{sun2022paco}.
Furthermore, the issue of gradient conflict is exacerbated by an increase in the number of tasks (detailed in Section \ref{sec:rethink}), underscoring the urgency for effective solutions to these challenges. 

\begin{table}[t!]
\centering
\caption{Summary of existing methods from the perspectives of awareness of multi-task training (MT), harmonious gradients (HG), and task identification (TI).  \textcolor[RGB]{34,139,34}{\Checkmark} and \textcolor{red}{\XSolidBrush} represent whether the method possesses the specific feature or not.}
\label{tab:related}
\small
\centering
\begin{tabular}{l|ccc}
\toprule[2pt]  Research work & MT aware& HG aware & TI aware \\ \midrule
DT \citep{DT} & \textcolor{red}{\XSolidBrush}& \textcolor{red}{\XSolidBrush}& \textcolor{red}{\XSolidBrush}\\
PromptDT \citep{PDT} & \textcolor{red}{\XSolidBrush}& \textcolor{red}{\XSolidBrush}& \textcolor{red}{\XSolidBrush}\\
MGDT \citep{lee2022multi}& \textcolor[RGB]{34,139,34}{\Checkmark}& \textcolor{red}{\XSolidBrush}& \textcolor{red}{\XSolidBrush}\\
MTDIFF \citep{he2023diffusion} & \textcolor[RGB]{34,139,34}{\Checkmark}& \textcolor{red}{\XSolidBrush}& \textcolor{red}{\XSolidBrush}\\
HarmoDT (Ours) & \textcolor[RGB]{34,139,34}{\Checkmark}& \textcolor[RGB]{34,139,34}{\Checkmark}& \textcolor{red}{\XSolidBrush}\\
G-HarmoDT~(Ours)& \textcolor[RGB]{34,139,34}{\Checkmark}& \textcolor[RGB]{34,139,34}{\Checkmark}& \textcolor[RGB]{34,139,34}{\Checkmark}\\
\bottomrule[2pt]
\end{tabular}
\end{table}

Existing works on offline MTRL generally address the problem in one of three ways \citep{sun2022paco}: 
1) developing shared structures for the sub-policies of different tasks, as explored in works by \citet{calandriello2014sparse, yang2020multi, lin2022switch}; 
2) optimizing task-specific representations to condition the policies, as discussed by \citet{sodhani2021multi, lee2022multi, he2023diffusion};
3) addressing the conflicting gradients arising from different task losses during training, a focus of research by \citet{yu2020gradient, chen2020just, liu2021conflict}.
While these methods have demonstrated effectiveness in different scenarios, they often fall short of adequately addressing the occurrence of conflicting gradients that stem from indiscriminate parameter sharing \citep{guangyuan2022recon}.
Additionally, many approaches require task identifiers, as summarized in Table~\ref{tab:related}, to determine the relevant parameter subspace or task-specific representations during inference~\citep{sun2022paco, yang2020multi, he2023diffusion}. 
This reliance limits their applicability in real-world environments where task content frequently varies and the current task is usually unknown.
As shown in Figure~\ref{fig:performance_num}, as the number of tasks increases and task identifiers are absent, existing MTRL algorithms experience significant performance degradation.

\begin{figure}[t!]
\centering 
\subfigure[Conflicting during MTRL.]{
\label{fig:intro_4}
\hspace{-15pt}
\includegraphics[width=0.245\textwidth]{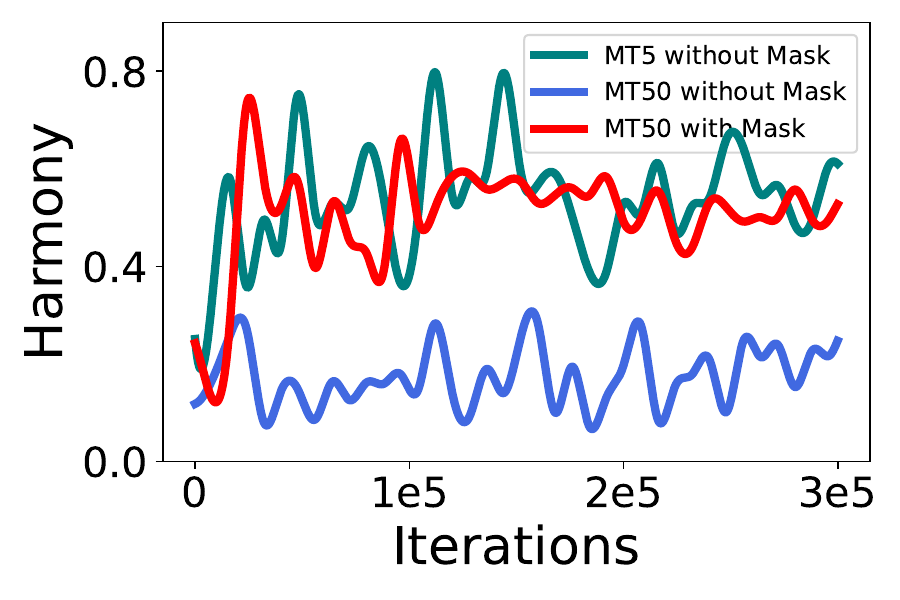}}\!\!\!
\hspace{15pt}
\subfigure[Success with task masks.]{
\hspace{-15pt}
\label{fig:intro_5}
\includegraphics[width=0.21\textwidth]{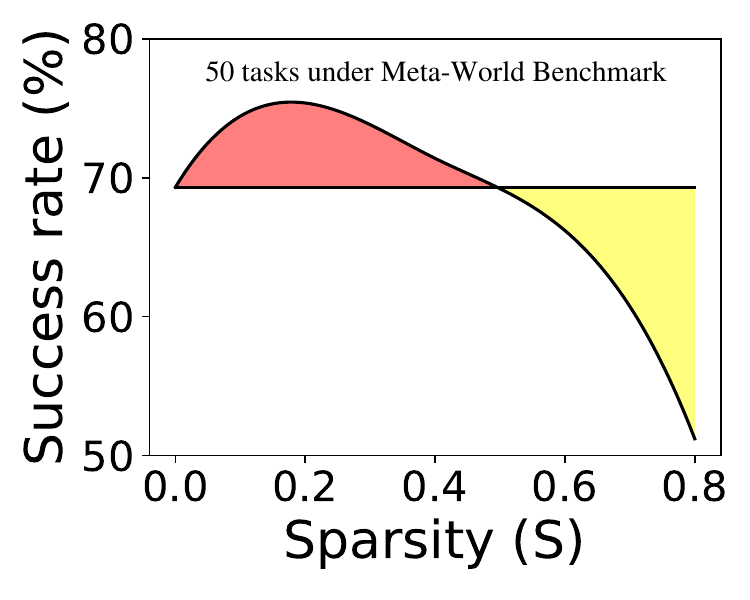}}
\caption{Illustration of the averaged harmony score among trainable weights during training for policies with and without randomly initialized masks (left panel), and success rates with varying mask sparsity levels (right panel) in the Meta-World benchmark. The averaged harmony score, defined in Section~\ref{sec:conflict}, reflects better harmony with higher values.}
\label{fig:redun_conflict}
\vspace{-0.2cm}
\end{figure}

To reduce the occurrence of the conflicting gradient, the idea of adopting distinct parameter subspaces for each task is straightforward.
Empirical observations, depicted by Figure~\ref{fig:intro_4}, affirm that the application of masks significantly mitigates conflicts, leading to considerable performance gains across various sparsity ratios\footnote{Sparsity ratio refers to the percentage of inactive weights.}, as contrasted with the non-mask baseline shown in Figure~\ref{fig:intro_5}.
Building upon these insights, we propose Harmony Multi-Task Decision Transformer (HarmoDT) to identify an optimal harmony subspace of parameters for each task by incorporating trainable task-specific masks during MTRL training. 
We model it as a bi-level optimization problem, employing a meta-learning framework to discern the harmony subspace mask via gradient-based techniques.
At the upper level, we focus on learning mask that delineates the harmony subspace, while at the inner level, we update parameters to augment the collective performance of the unified model under the guidance of the mask.
To further eliminate the need for task identifiers, we design a group-wise variant, G-HarmoDT, which clusters similar tasks into groups based on their gradient information and employs a gating module to infer the identifiers of these groups.
Empirical evaluations of HarmoDT and its variants across diverse tasks, with and without task identifiers, demonstrate significant improvements over state-of-the-art algorithms: 8\% in task-provided settings, 5\% in task-agnostic settings, and 10\% in unseen tasks.
Additionally, we provide extensive ablation studies on various aspects, including scalability, model size, hyper-parameters, and visualizations, to comprehensively validate our approach.

In summary, our research makes three significant contributions to the field of MTRL:
\begin{itemize}[leftmargin=15pt]
    \item 
    We revisit MTRL challenges from the perspective of sequence modeling, analyzing issues such as gradient conflicts that intensify with an increasing number of tasks, as well as the reliance on task identifiers during inference, which limits applicability in real-world scenarios where task content frequently varies and the current task is often unknown~(Section \ref{sec:rethink}).
    \item To address the issue of conflicting gradients, we introduce a novel solution, HarmoDT, which seeks to identify an optimal harmonious subspace of parameters. This is formulated as a bi-level optimization problem, approached through meta-learning and solved using gradient-based methods.~(Section \ref{sec:method}).
    \item To eliminate reliance on task identifiers, we design a group-wise variant, G-HarmoDT, which clusters similar tasks into groups based on their gradient information and employs a gating module to infer the identifiers of these groups (Section \ref{sec:method}).
    \item We demonstrate the superior performance of our HarmoDT and G-HarmoDT through rigorous testing on a broad spectrum of benchmarks, establishing its state-of-the-art effectiveness in MTRL scenarios (Section \ref{sec:exp}).
\end{itemize}

\section{Related Work}
\subsection{Offline Reinforcement Learning}
In offline RL, unlike the fundamentally online paradigm \citep{sutton2018reinforcement}, learning is performed without interaction with the environment \citep{levine2020offline}.
Instead, it depends on a dataset $\mathcal{D}$ comprising trajectories generated from various behavior policies.
The objective of offline RL is to utilize this dataset $\mathcal{D}$ to learn a policy that maximizes the expected return.
Previous studies have employed constrained or regularized dynamic programming techniques to minimize deviations from the behavior policy \citep{TD3BC, CQL, IQL}.
Another prevalent method is Return-conditioned Behavior Cloning (BC), which involves learning the state-to-action mapping through supervised learning from the dataset. 
Transformer-based policies \citep{TRL} have been influential in return-conditioned BC, predicting subsequent actions based on sequences of past experiences, including state-action-reward triplets. 
This approach inherently restricts the learned policy within the behavior policy's bounds and focuses on conditioning the policy on specific metrics for future trajectories \citep{DT, TT, GDT, QT}.
Recently, there has been increasing interest in incorporating diffusion models into offline RL. 
This alternative decision-making strategy leverages the success of generative modeling, providing potential improvements for addressing offline RL challenges. 
Diffuser and its variants \citep{Diffuser, DD, chen2022offline} employ diffusion-based generative models to represent policies or model dynamics, achieving superior performance across various tasks.

\subsection{Multi-Task Learning}
In the context of multi-task learning, several methodologies have been developed to mitigate the effects of conflicting gradients.
PCGrad \citep{yu2020gradient} projects each task's gradient onto the orthogonal plane of another's, subsequently updating parameters using the mean of these projected gradients. 
Graddrop \citep{chen2020just} employs a stochastic approach, randomly omitting certain gradient elements based on their conflict intensity. 
CAGrad \citep{liu2021conflict} manipulates gradients to converge towards a minimum average loss across tasks.
Instead of adjusting gradients post hoc as in these methods, we proactively utilize gradient data to inform the selective activation of parameters for tasks through a masking mechanism. 
This direct intervention at the parameter level allows the model to update without the typical interferences found in gradient-level adjustments, fostering a more streamlined and potentially more efficacious optimization process.
For comparison, we integrate classical methods PCGrad and CAGrad with MTDT in our experiments.

\subsection{Multi-Task Reinforcement Learning}
Multi-task RL aims to learn a shared policy for a diverse set of tasks, with numerous approaches proposed in the literature \citep{yang2020multi, sarafian2021recomposing, sodhani2021multi}. 
By leveraging the expressive scalability and parameter-sharing capabilities of Transformer architectures alongside the sequence modeling paradigm in offline RL, many works aim to design a highly generalist Transformer-based policy \citep{Gato, lee2022multi} capable of handling multiple diverse tasks.
A straightforward approach involves pooling extensive datasets collected from various scenarios to address multiple tasks simultaneously. This pooled data enables the model to learn strategies more effectively across all tasks compared to training each task in isolation.
While sharing data across all tasks is expected to improve performance by exploiting task similarities through parameter sharing, this approach can lead to serious conflicting gradients due to indiscriminate parameter sharing \citep{guangyuan2022recon}, resulting in sub-optimal performance.
Moreover, current MTRL methods often require access to the task identifier to obtain specific information about the current task as summarized in Table~\ref{tab:related} \citep{PDT, sun2022paco, he2023diffusion}, which can be problematic in realistic applications.
To reduce conflicting gradients, we propose Harmony Multi-Task Decision Transformer (HarmoDT) \citep{HarmoDT}, which identifies an optimal harmony subspace of parameters for each task by using trainable task-specific masks during MTRL training. 
To further eliminate the need for task identifiers, we design a group-wise variant, G-HarmoDT, which clusters similar tasks into groups based on their gradient information and employs a gating module to infer the identifiers of these groups.
\section{Rethinking Sequence Modeling with MTRL}
\label{sec:rethink}

Recent works in offline RL conceptualize it as sequence modeling (SM), effectively transforming extensive datasets into potent decision-making systems. 
This approach is advantageous for multi-task RL, offering a high-capacity model that accommodates task discrepancies and assimilates comprehensive knowledge from diverse datasets. 
However, the direct application of such high-capacity sequential models to multi-task RL introduces significant algorithmic challenges. 
In this section, we outline the primary challenges, including conflicting gradients and the dependence on task identifiers. We further explore the concepts of parameter subspace and harmony, laying the groundwork for the motivation.

\subsection{Conflicting Gradients} 
\label{sec:conflict}

We first investigate conflicting gradients. 
In a multi-task training context, the aggregate gradient, $\hat{\rvg}$, is computed across multiple tasks and is defined as
\begin{equation}
    \hat{\rvg}= \E_{\gT_i \sim p(\gT)} \nabla \gL_{\gT_i}(\theta)  =\frac{1}{N}\sum_{i=1}^N \rvg_i(\theta),
\end{equation}
where $\theta$ represents the trainable parameter vector and $\rvg_i$ is the gradient vector for task $\gT_i$.
In scenarios where tasks are diverse, the gradients $\rvg_i$ from different tasks may conflict significantly, a phenomenon known as gradient conflicts and negative transfer in multi-task learning \citep{guangyuan2022recon, tang2023concrete}.

\begin{definition}[Harmony Score on a Single Weight]
    The harmony score of the j-th element in the weights vector is estimated by calculating the corresponding coordinate of the element-wise product of the task gradient and the total gradient, denoted as $(\rvg_i \odot \hat{\rvg})_j=\rvg_{i,j} \times \hat{\rvg}_j$.
\end{definition}

\begin{definition}[Averaged Harmony Score]
    The overall harmony score across all weights is evaluated using $\frac{1}{NK}\sum_{i=1}^{N}\sum_{j=1}^{|\theta|} \frac{(\rvg_i \odot \hat{\rvg})_j}{|\rvg_{i,j}| |\hat{\rvg}_j|}$, where $|\theta|$ and $N$ denote the number of weights and tasks. This score, ranging between -1 and 1, reflects the degree of alignment among tasks.
\end{definition}

To substantiate the presence of conflicts in MTRL, we establish two metrics to measure harmony score in weights among tasks and conduct experiments utilizing the Prompt-DT method on 5 and 50 tasks from the Meta-world benchmark, recording the average harmony score.
As illustrated in Figure~\ref{fig:intro_4}, the averaged harmony score significantly diminishes with the escalation in the number of tasks, indicating pronounced conflicts among tasks and underscoring the imperative to address these conflicts in MTRL.

\subsection{Parameter Subspace and Harmony} 
\label{sec:redun}
Parameter subspace, a concept prevalent in pruning-aware training \citep{pruning}, aims to maintain comparable performance while achieving a sparse model. 
In the context of MTRL via SM, pruning to preserve distinct parameter subspaces for each task markedly alleviates gradient conflicts.
To validate this, we conduct experiments on 50 tasks from the Meta-World benchmark. 
Each task $\gT_i$ is assigned a randomly initialized mask $\mM^{\gT_i}$ with a specific sparsity ratio $\mathrm{S}$. 
During training, this mask modulates both the trainable parameters and gradients as
\begin{equation}
\label{eq:mask_gradient}
    \bar{\rvg}_i =  \nabla \gL_{\gT_i}(\theta \odot \mM^{\gT_i}) \odot \mM^{\gT_i}, ~~ i = 1, 2, \dots, N,
\end{equation}
where $\odot$ represents element-wise multiplication, and $\mM^{\gT_i}$ is a binary vector.
Intriguingly, as shown in Figure~\ref{fig:intro_5}, applying the mask could result in enhanced performance across a wide range of sparsity ratios. 
This improvement, coupled with the significantly higher harmony score in multi-task settings shown in Figure~\ref{fig:intro_4}, suggests that maintaining a subspace of parameters with task-specific masks effectively reduces conflicts arising from unregulated parameter sharing.

\subsection{Gating Network}
In multi-task training, various approaches \citep{he2023diffusion, PDT,lee2022multi} aim to leverage the scalability and parameter-sharing capabilities of Transformer-based policies.
However, these methods often require a task identifier during inference to access metadata about the current task. 
As illustrated in Figure \ref{fig:performance_num}, applying these methods in a task-agnostic setting—where information about the current task is unavailable—leads to a notable performance decline.

\begin{figure}[!t]
\centering
\includegraphics[width=0.48\textwidth]{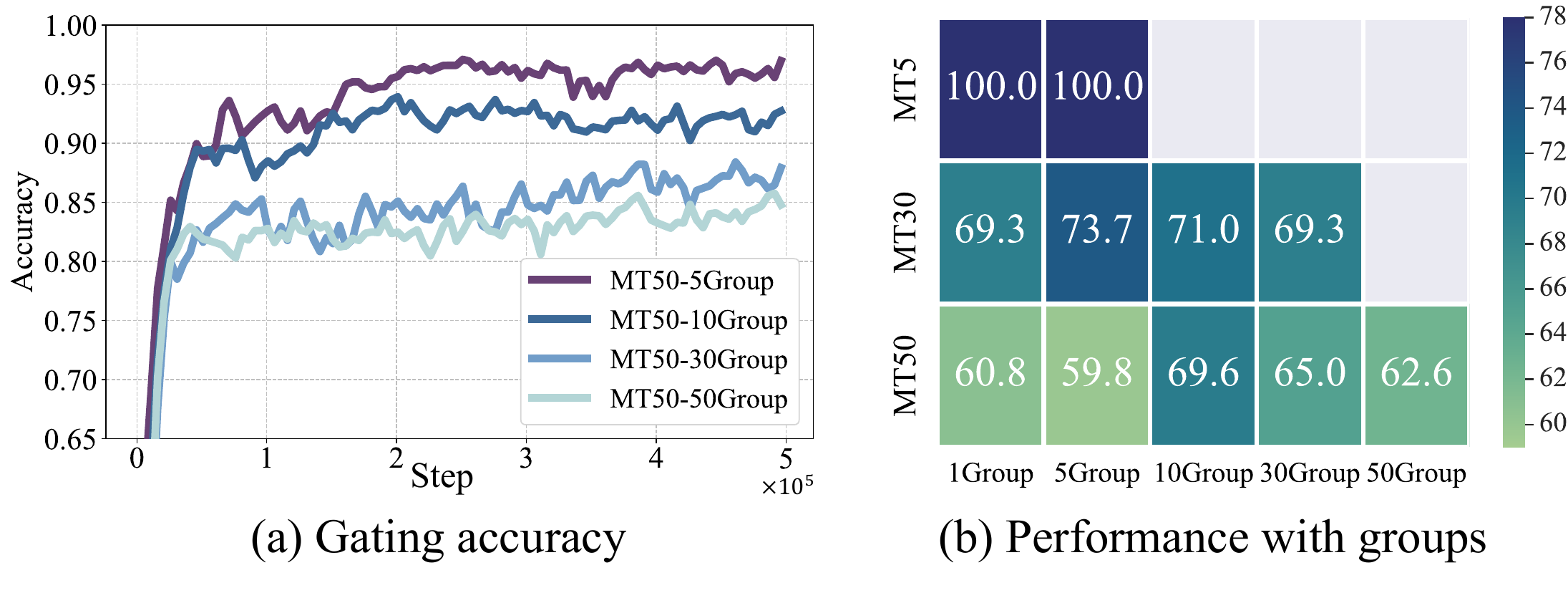}
\caption{(a) Accuracy of the gating module in classifying tasks into groups under 50 tasks of Meta-World (e.g., 5 groups mean each group contains 10 tasks). The accuracy of the gating network decreases significantly as the number of groups increases. 
(b) Performance in task-agnostic settings with different numbers of groups and the corresponding gating module under Meta-World benchmark. Directly combining the gating network with HarmoDT results in sub-optimal performance. For example, in MT50 with 50 tasks, 10Group (G-HarmoDT) achieves the best performance instead of 50Group (HarmoDT).
An in-depth analysis is provided in Section~\ref{sec:exp}.
}
\label{fig:reduce_burden}
\end{figure}

Gating networks are frequently employed in deep learning to differentiate between various inputs. 
To address the need for task identification in task-agnostic settings, a straightforward approach is to train a simple gating network in combination with these masking methods.
However, as shown in Figure \ref{fig:reduce_burden}{\color{red}(a)}, the accuracy of the gating network significantly declines as the number of groups increases. 
This reduction in accuracy results in sub-optimal performance when the gating network is directly combined with HarmoDT, which requires precise task differentiation, as illustrated in Figure \ref{fig:reduce_burden}{\color{red}(b)}.
This observation motivates the clustering of similar tasks. Clustering not only improves the accuracy of the gating network but also reduces the learning burden associated with optimal mask learning.

\subsection{Motivation} 
The complexity of MTRL increases significantly with more tasks, mainly due to escalating gradient conflicts. 
This challenge arises from unregulated parameter sharing, which aims to exploit task similarities but often results in performance degradation.
In response to these challenges, task-specific masks have been proposed to maintain distinct parameter subspaces for each task, preventing one task's learning from negatively impacting others.
While this strategy represents a step towards mitigating gradient conflicts, it introduces new challenges: 1) determining optimal masks, and 2) the reliance on task identifiers.
The complexity of optimizing masks is compounded by the dynamic and often non-linear task interactions within the shared model space. 
Additionally, the accuracy of the gating network declines with an increasing number of tasks, as distinguishing between similar tasks becomes progressively more challenging.

To determine optimal masks that balance shared learning and task-specific adaptation, we propose HarmoDT, a novel solution using a bi-level optimization strategy within a meta-learning framework.
This approach leverages gradient-based techniques to meticulously explore and exploit the parameter space and identify a harmony parameter subspace for tasks, optimizing overall MTRL performance.
To eliminate dependency on task identifiers, we introduce a group-wise variant, G-HarmoDT, which assigns similar tasks a shared group identifier, thereby reducing the learning burden on the gating network. 
This formulation reframes the problem as identifying an optimal "sweet point" that balances the performance of the gating network with the management of conflicting gradients among tasks within each group.

\section{Method}\label{sec:method}
\subsection{Preliminaries}
\subsubsection{Offline Reinforcement Learning}
RL aims to learn a policy $\pi_{\theta}(\rva | \rvs)$ maximizing the expected cumulative discounted rewards $\E[\sum_{t=0}^{\infty} \gamma^t \gR(\rvs_t, \rva_t)]$ in a Markov decision process (MDP), which is a six-tuple $(\gS, \gA, \gP, \gR, \gamma, d_0)$, with state space $\gS$, action space $\gA$, environment dynamics $\gP(\rvs' | \rvs, \rva): \gS \times \gS \times \gA \rightarrow [0,1]$, reward function $\gR: \gS \times \gA \rightarrow \R$, discount factor $\gamma \in [0, 1)$, and initial state distribution $d_0$ \citep{sutton2018reinforcement}.
The action-value or Q-value of a policy $\pi$ is defined as $Q^{\pi}(\rvs_t, \rva_t) = \E_{\rva_{t+1}, \rva_{t+2} \dots \sim \pi} [\sum_{i=0}^{\infty} \gamma^i \gR(\rvs_{t+i}, \rva_{t+i})]$.
In offline RL \citep{levine2020offline}, a static dataset $\gD = \{(\rvs, \rva, \rvs', r)\}$, collected by a behavior policy $\pi_{\beta}$, is provided and algorithms learn a policy entirely from this static dataset without any online interaction with the environment.

\subsubsection{Multi-task Reinforcement Learning}
In the multi-task setting, different tasks can have different reward functions, state spaces, and transition functions. 
We consider all tasks to share the same action space with the same embodied agent.
Given a specific task $\gT \sim p(\gT)$, a task-specified MDP can be defined as $(\gS^{\gT}, \gA^{\gT}, \gP^{\gT}, \gR^{\gT}, \gamma, d_0^{\gT})$.
Instead of solving a single MDP, the goal of multi-task RL is to find an optimal policy that maximizes expected return over all the tasks: $\pi^* = \argmax_{\pi} \E_{\gT \sim p(\gT)} \E_{\rva_t \sim \pi} [\sum_{t=0}^{\infty} \gamma^t r_t^{\gT}] $.
The static dataset $\gD$ correspondingly is partitioned into per-task subsets as $\gD = \cup_{i=1}^N \gD_i$, where $N$ is the number of tasks.

\subsubsection{Prompt Decision Transformer}
The integration of Transformer \citep{transformer} architecture into offline RL for sequential modeling (SM) has gained prominence in recent years. 
NLP studies show that Transformers pre-trained on large datasets exhibit strong few-shot or zero-shot learning abilities within a prompt-based framework \citep{liu2023pre, brown2020language}. 
Building on this, Prompt-DT adapts the prompt-based approach to offline RL, using trajectory as prompts, consisting of state, action, and return-to-go tuples ($\rvs^*$, $\rva^*$, $\hat{r}^*$) to provide directed guidance for RL agents. 
Each element marked with $\cdot^*$ relates to the trajectory prompt. 
%
%
During training with offline collected data, Prompt-DT utilizes $\tau_{i,t}^{input}=(\tau_i^*, \tau_{i,t})$ as input for each task $\gT_i$. 
Here, $\tau^{input}_{i,t}$ consists of the $K^*$-step trajectory prompt $\tau_i^*$ and the most recent $K$-step history $\tau_{i,t}$, and is formulated as follows:
\begin{align}
\label{eq:input}
    &\tau^{input}_{i, t} = (\hat{r}^*_{i, 1}, \rvs^*_{i, 1}, \rva^*_{i, 1}, \dots,  \hat{r}^*_{i, K^*}, \rvs^*_{i, K^*}, \rva^*_{i, K^*}, \nonumber \\ 
    &\quad\hat{r}_{i, t-K+1}, \rvs_{i, t-K+1}, \rva_{i, t-K+1}, \dots,  \hat{r}_{i, t}, \rvs_{i, t}, \rva_{i, t}).
\end{align}
The prediction head connected to the state token $\rvs$ is designed to predict the corresponding action $\rva$. 
For continuous action spaces, training objective minimizes the mean-squared loss as
\begin{equation}
\label{eq:dtloss}
   \gL_{DT}\! =\! \mathbb{E}_{\tau^{input}_{i, t} \sim \gD_i} \left[ \frac{1}{K}\!\! \sum_{m=t-K+1}^t\!\! (\rva_{i, m} - \pi(\tau^*_i, \tau_{i,m} ) )^2 \right]. 
\end{equation}

\begin{figure*}[t!]
\centering
\includegraphics[width=0.98\textwidth]{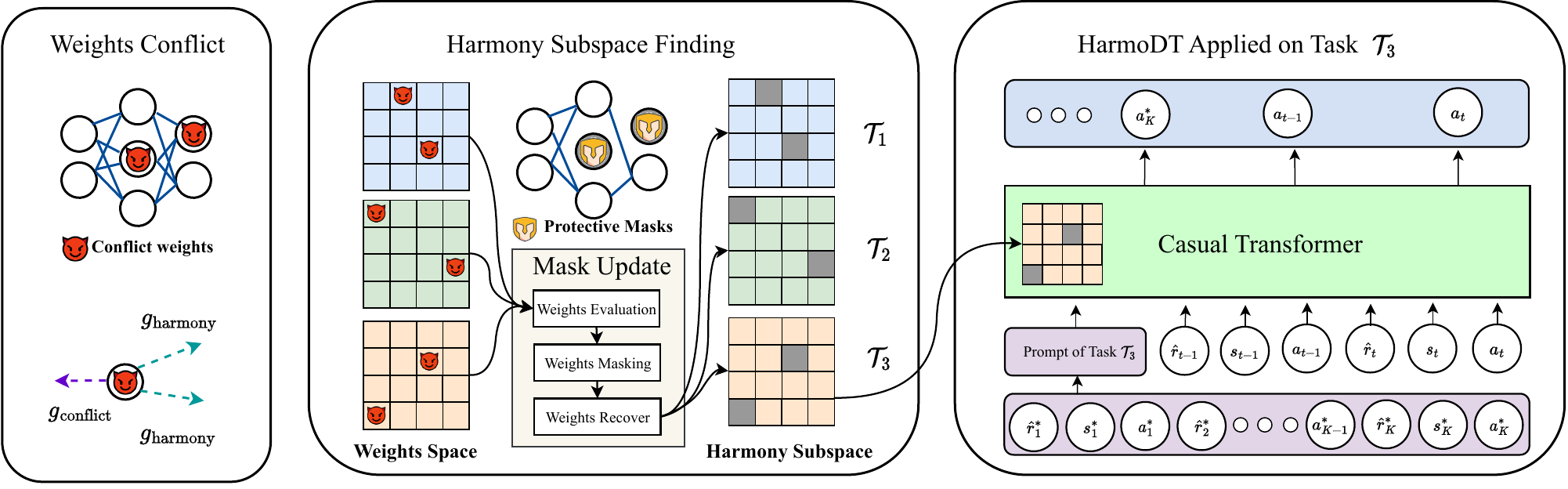}
\vspace{-0.2cm}
\caption{Illustration of the conflicting problem and the framework of HarmoDT to find a harmony subspace for each task. 
The left panel shows the conflicting phenomenon reflected by divergent task-specific gradients.
The middle panel illustrates the procedure to find a harmony subspace for each task via the mask learning.
The right panel demonstrates the workflow of HarmoDT based on the DT architecture with prompts when handling a task, such as $\gT_3$.}
\label{fig:pipeline}
\vspace{-0.3cm}
\end{figure*}
\subsection{HarmoDT: Find optimal subspace}
To address the aforementioned problem, we introduce a meta-learning framework that discerns an optimal harmony subspace of parameters for each task, enhancing parameter sharing and mitigating gradient conflicts. 
This problem is formulated as a bi-level optimization, where we meta-learn task-specific masks to define the harmony subspace. 
Mathematically, we can express the problem as:
\begin{align}
    \max_{\sM} ~ & ~ \E_{\gT_i \sim p(\gT)} [\sum_{t=0}^{\infty} \gamma^t \gR^{\gT_i}(s_t, \pi (\tau_{i,t}^{input} | \theta^{*\gT_i}))], \\
    \text{s.t.} ~~ & \theta^* = \argmin_{\theta} \E_{\gT_i \sim p(\gT)} \gL_{DT} (\theta, \sM), \\
    \text{where}& ~ \theta^{*\gT_i}=\theta^* \odot \mM^{\gT_i}, \sM = \{ \mM^{\gT_i} \}_{\gT_i \sim p(\gT)},
\end{align}
where $\mM^{\gT_i}$ represents a binary task mask vector corresponding to $\gT_i$, and $\sM$ denotes the set of all task masks.
The goal at the upper level is to learn a task-specific mask that identifies the harmony subspace for each task. 
Concurrently, at the inner level, the objective is to optimize the algorithmic parameters $\theta$, maximizing the collective performance of the unified model under the guidance of the task-specific masks.
The framework for our harmony subspace learning is depicted in Figure \ref{fig:pipeline}.
Subsequent sections are meticulously dedicated to elucidating the methodology for selecting the harmony subspace, detailing the metrics for evaluating the importance and conflicts of weights, the sophisticated update mechanism for task masks, and delineating the procedural intricacies of the algorithm.

\begin{algorithm}[t!]
    {\footnotesize
    \caption{HarmoDT}
    \label{alg:harmodt}  
    \begin{algorithmic}
    \STATE {\bfseries Input:} A set of tasks $\sT = \{ \gT_1,\dots,\gT_N \}$, maximum iteration $E$, episode length $T$, target return $\tG^*$, learning rate $\eta$, hyper-parameters $\{\eta_{\max}, \eta_{\min}, t_m, \lambda, \mathrm{S}, \text{thresh} \}$.
    \STATE {\color{gray}\te{// initializing stage}} \\
    \STATE Initialize the parameters of the network $\theta_0$ , the set of task masks  $\sM = \{\mM^{\gT_1},\dots, \mM^{\gT_N}\}$ through ERK with $\mathrm{S}$, and $t\leftarrow1$.
    \STATE {\color{gray}\te{// training stage}} \\
    \WHILE{$t \leq E$}
        \STATE $\alpha_t=\lceil\eta_{max}+\frac{1}{2}(\eta_{min}-\eta_{max})(1+\cos{(2\pi \frac{t}{E})})\rceil$.
        \STATE \colorbox{gray!20}{$\sM$=\textbf{Mask\_Update}($\sT, \sM, \theta_t, \alpha_t, \lambda$).}

        \WHILE{$t \bmod t_m \neq 0$}
            \STATE sample a task $\gT_i$ from $\sT$, and a batch of data $\tau_i$ from the dataset$\gD_i$. 
            \STATE $\theta \leftarrow \theta - \eta \nabla \gL_{\gT_i}(\theta_t \odot \mM^{\gT_i}) \odot \mM^{\gT_i}$.
            \STATE $t \leftarrow t + 1$.
        \ENDWHILE
        \STATE $t \leftarrow t+1$.
    \ENDWHILE
    \STATE {\color{gray}\te{// inference stage}} \\
    \FOR{$i = 1, \dots, N$}
        \STATE Initialize history $\tau$ with zeros, desired reward $\hat{r} = \tG_i^*$, prompt $\tau^*$, the parameters $\theta_i \leftarrow \theta \odot \mM^{\gT_i}$, and $j\leftarrow1$.
        \FOR{$j \leq T$}
            \STATE Get action $a = \text{Transformer}_{\theta_i}(\tau^*, \tau)[-1]$.
            \STATE Step env. and get feedback $\rvs, \rva, r, \hat{r} \leftarrow \hat{r} - r$.
            \STATE Append $[\rvs, \rva, \hat{r}]$ to recent history $\tau$.
            \STATE $j\leftarrow j+1$.
        \ENDFOR
    \ENDFOR
    \end{algorithmic}
 }
\end{algorithm}
\subsubsection{Weights Evaluation}\label{sec:harmo_metric}
During training, our aim is to iteratively identify a harmony subspace for each task by assessing trainable parameter conflicts and importance. 
This involves defining two metrics: the Agreement Score and the Importance Score, to gauge the concordance and significance of weights.

\begin{definition}[Agreement Score]
For each task $\gT_i$ with a set of task masks $\sM$, the agreement score vector of all trainable weights is defined as follows: $A(\gT_i)=\bar{\rvg}_{i}\cdot \frac{1}{N} \sum_{j=1}^N \bar{\rvg}_{j}$, where $\bar{\rvg}_{j}$ denotes the masked gradients for task $\gT_j$, which is defined in Equation \ref{eq:mask_gradient}.
\end{definition}
\begin{definition}[Importance Score]
We measure the significance of parameters for task $\gT_i$ either through the absolute value of the parameters $I_M(\gT_i)=|(\theta\odot\mM^{\gT_i})|$, indicating magnitude-based importance, or through the Fisher information $I_F(\gT_i)= \left(\nabla \log \gL_{\gT_i}\left( \theta\odot\mM^{\gT_i}\right) \odot \mM^{\gT_i} \right)^2$, reflecting the parameters' impact on output variability.
\end{definition}

For task $\gT_i$, $A(\gT_i)$ reflects the gradient similarity between the task-specific and the average masked gradients, while $I_M(\gT_i)_j$ and $I_F(\gT_i)_j$ measure the j-th element's importance. 
Lower values of $A(\gT_i)_j$ or $I_{M/F}(\gT_i)_j$ indicate increased conflict or diminished importance regarding the j-th element of the trainable parameters for the respective task.
The Harmony Score $H(\gT_i, \mM^{\gT_i})_j$ for the j-th parameter of task $\gT_i$ is computed as a weighted balance between the Agreement and Importance Scores, moderated by a balance factor $\lambda$:
\begin{equation} \nonumber
    \label{eq:init}
    \!\!H(\gT_i,\mM^{\gT_i})_j=\left\{
    \begin{array}{ll}
    \!\!A(\gT_i)_j+\lambda I_{M/F}(\gT_i)_j, & (\mM^{\gT_i})_j =1, \\
    \!\! \inf ,  & (\mM^{\gT_i})_j =0.\\
    \end{array} \right. 
\end{equation}
Parameters that have been already masked (i.e., $(\mM^{\gT_i})_j = 0$) are assigned an infinite Harmony Score to prevent their re-selection due to the pre-existing conflicts.

\subsubsection{Mask Update}\label{sec:find}

For a given sparsity $\mathrm{S}$ and task masks $\sM$, we periodically assess the harmony subspace of trainable weights $\theta$ across all tasks. 
This process involves masking\footnote{When the term "mask" is used as a verb, it refers to the action of rendering the corresponding parameter inactive or modifying the mask vector by changing the value from 1 to 0.} $\alpha_t$ of the most conflicting parameters and subsequently recovering an equal number of previously masked parameters that have transitioned to harmony after the subsequent iterative training process. 
As delineated in Algorithm~\ref{alg:harmodt_update}, this procedure encompasses three key steps: Weights Evaluation, Weights Masking, and Weights Recovery.

\begin{algorithm}[t!]
    {\footnotesize
    \caption{Mask Update}
    \label{alg:harmodt_update}
    \begin{algorithmic}
    \STATE {\bfseries Input:} A set of tasks $\sT$, a set of task masks $\sM$, trainable weights vector $\theta_t$, updating number $\alpha_t$, and hyper-parameters $\{ \lambda \}$.
    \FOR{$i= 1,\dots,N$}
        \STATE $\rvg_i= \nabla \gL_{\gT_i} (\theta) $.
        \STATE $\bar{\rvg}_i= \nabla \gL_{\gT_i} (\theta \odot \mM^{\gT_i}) \odot \mM^{\gT_i}$.
    \ENDFOR
    \FOR{$i= 1,\dots,N$}
        \STATE {\color{gray}\te{// Weights Evaluation}} \\
        \STATE Calculate $H(\gT_i, \mM^{\gT_i})$ with $\lambda$, $\rvg_i$ and $\bar{\rvg}$ as Sec.~\ref{sec:harmo_metric}.
        \STATE {\color{gray}\te{// Weights Masking}} \\
        \STATE $\mM^{\gT_i}=\mM^{\gT_i}-\operatorname{ArgBtmK}_{\alpha_t}\left(H(\gT_i, \mM^{\gT_i})\right).$
        \STATE {\color{gray}\te{// Weights Recovery}}
        \STATE $\mM^{\gT_i}=\mM^{\gT_i}+\operatorname{ArgTopK}_{\alpha_t}(\rvg_i \odot \frac{1}{N} \sum_{j=1}^N \bar{\rvg}_j -\inf \odot \mM^{\gT_i}).$
    \ENDFOR
    \end{algorithmic}
{\textbf{Output}:$\sM=\{\mM^{\gT}\}.$}
}
\end{algorithm}

\paragraph{Weights Masking}
Employing the Harmony Score, we identify and mask the most conflicting and less significant weights within the harmony subspace as follows:
\begin{equation}
\label{eq:weightmask}
    \mM^{\gT_i}=\mM^{\gT_i}-\operatorname{ArgBtmK}_{\alpha_t}\left(H(\gT_i,\mM^{\gT_i})\right),
\end{equation}
where $\alpha_t$ represents the number of masks altered in the t-th iteration, and $\operatorname{ArgBtmK}_{\alpha_t}(\cdot)$ generates zero vectors matching the dimension of $\mM^{\gT_i}$, marking the positions of the top-$\alpha_t$ smallest values from $H(\gT_i,\mM^{\gT_i})$ with 1.

\paragraph{Weights Recover}
To maintain a fixed sparsity ratio and recover weights that have transitioned from conflict to harmony, the following recovery process is applied:
\begin{equation}\label{eq:weightrecover}
\mM^{\gT_i}=\mM^{\gT_i}+\operatorname{ArgTopK}_{\alpha_t}(\rvg_i \odot \frac{1}{N}\sum_{j=1}^N \bar{\rvg}_j-\inf \odot \mM^{\gT_i}),
\end{equation}
where $\operatorname{ArgTopK}_{\alpha_t}(\cdot)$ generates zero vectors and set the positions corresponding to the top-$\alpha_t$ largest values from $(\rvg_i \odot \frac{1}{N}\sum_{j=1}^N \bar{\rvg}_j-\inf \odot \mM^{\gT_i})$ to 1.
This step ensures the reintegration of previously conflicting weights that have harmonized after subsequent iterations and the term $(-\inf \odot \mM^{\gT_i})$  prevents the re-selection of active parameters.  

\subsubsection{Total Framework}
\label{sec:inferofharmo}
Algorithm~\ref{alg:harmodt} provides the meta-learning process for the task mask set $\sM$ and the update mechanism for the trainable parameters $\theta$ of the unified model.
Given a set of source tasks, we first initialize corresponding masks through the ERK technique \citep{ERK} with a predefined sparsity ratio $\mathrm{S}$ for each task (See Appendix C for more details).
In the inner loop, we optimize the parameters of the unified model under the guidance of the task-specific mask:
\begin{equation}
    \theta_{t+1}=\theta_{t}-\eta \E_{\gT_i \sim p(\gT)} \nabla \gL_{\gT_i}(\theta \odot \mM^{\gT_i}) \odot \mM^{\gT_i}.
\end{equation}
Then, in the outer loop, we optimize task masks through the procedure detailed in Section \ref{sec:find} to find the harmony subspace for each separate task.
Considering the stability and efficiency of the updating, we adopt a cosine annealing strategy~\citep{cos} to control the updating number $\alpha_t$. Given the maximum iterations $E$, $\alpha_t$ in t-th iteration is defined as:
\begin{equation}
    \alpha_t=\lceil\eta_{max}+\frac{1}{2}(\eta_{min}-\eta_{max})(1+\cos{(2\pi \frac{t}{E})})\rceil,
\end{equation}
where $\lceil \cdot \rceil$ represents the round-up command, and $\eta_{min}$ and $\eta_{max}$ denote the lower and upper bounds, respectively, on the number of parameters that undergo changes during the mask update process.
In the inference stage, following \citep{he2023diffusion,PDT}, task IDs are provided on online environments and the task-specific mask is applied to the parameters for making decision. 

For unseen tasks that differ from training tasks but share identical states and transitions, we aggregate task-specific masks from training tasks to formulate a generalized model. 
The mask for unseen tasks is constructed as follows:
\begin{equation}
    \label{eq:unseen}
    \hat{\mM}_j=\left\{
    \begin{array}{ll}
0, & \sum_{i=1}^N \mM_{i,j} \leq \text{thresh},\\
1,  & \sum_{i=1}^N \mM_{i,j} > \text{thresh},\\
\end{array} \right. 
\end{equation}
where $\hat{M}$ denotes the mask for the unseen task, and `thresh' is a predefined threshold.

\begin{figure*}[t!]
\centering
\includegraphics[width=1.0\textwidth]{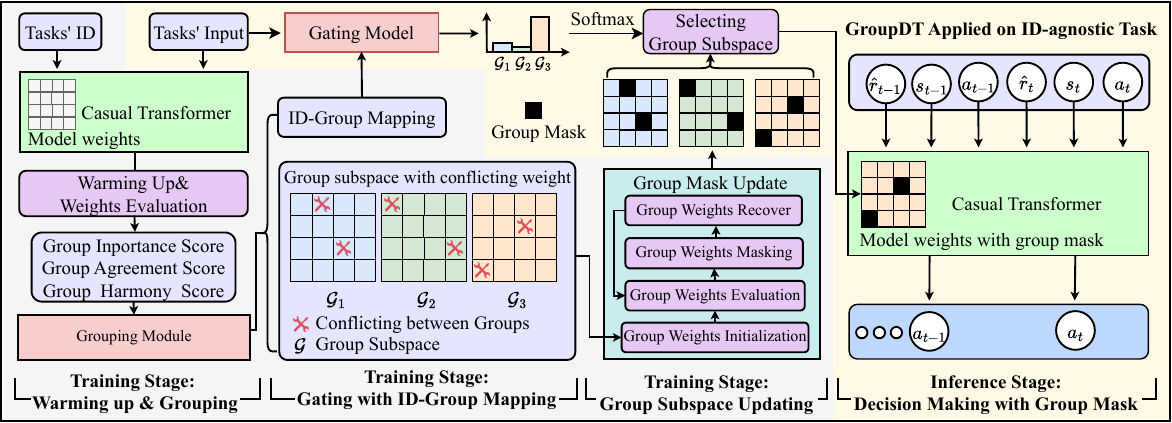}
\vspace{-0.4cm}
\caption{Overall framework of the G-HarmoDT: In the first stage, we warm up the weights and group tasks based on weight evaluations. In the second stage, we train a gating model using task inputs and ID-group mappings from the grouping module, while simultaneously updating the main model. In the third stage, we gradually update the group subspaces. During inference, the gating model provides the appropriate group ID and mask for the unknown task.}
\label{fig:framework}
\vspace{-0.3cm}
\end{figure*}
\subsection{G-HarmoDT: A Group-wise Variant}
To reduce the need for task identifier, we propose G-HarmoDT, a variant for group subspace training that partitions cohesive tasks into groups and assigns each group a harmonious subspace. 
During the inference stage, where task ID is unavailable, we employ the gating module to discern groups and determine the appropriate group subspace. 
Mathematically, we formulate the problem as follows:
\begin{align}
    \max_{\sM} ~ & ~ \E_{\gT_i \sim p(\gT)} [\sum_{t=0}^{\infty} \gamma^t \gR^{\gT_i}(s_t, \pi (\tau_{i,t}^{input} | \theta^{*\gG_j}))], \nonumber\\ 
    \text{s.t.} ~~ & \theta^* = \argmin_{\theta} \E_{\gT_i \sim p(\gT)} \gL_{DT} (\theta, \sM, \sG), \nonumber\\ 
    \text{where}& ~ \theta^{*\gG_j}=\theta^* \odot \mM^{\gG_j}, \sM = \{ \mM^{\gG_j}\}, \sG = \{ \gG_j \}, \nonumber
\end{align}
where $\sG$ denotes the set of all groups, $N^{G}$ is the number of groups defined as a hyper-parameter, $\gG_j$ represents a task group comprising a set of cohesive tasks, $\mM^{\gG_j}$ denotes the mask vector corresponding to $\gG_j$, and $\sM$ denotes the set of all masks.
To solve the objective, we modify the original HarmoDT and introduce key changes from three perspectives in the following sections: group-wise weight evaluation, group-wise mask update, and the gating module, followed by an overview of the entire framework.
%

\subsubsection{Group-wise Weights Evaluation}\label{sec:groupcalculate}
All metrics used to evaluate weights in HarmoDT (Section \ref{sec:harmo_metric}) are clustered at the group level, with the masked gradient for each task now derived based on the corresponding group mask:
\begin{equation}
    \label{eq:masked_gradient_group}
    \bar{\rvg}_i = \nabla \L_{\gT_i}(\theta\odot \mM^{\gG_j})\odot \mM^{\gG_j}, ~ \gT_i \in \gG_j.
\end{equation}

\begin{definition}[Group Agreement Score]
For a group $\gG_j\in\sG$ involving $n_j$ tasks and group masks $\mM^{\gG_j}\in\sM$, the group agreement score vector of all trainable weights is defined as $\text{GA}(\gG_j)= \sum_{\gT_i \in \gG_j} \bar{\rvg}_{i} \cdot \frac{1}{N^G} \sum_{\gG_k \in \sG} \sum_{\gT_i \in \gG_k}\bar{\rvg}_{i}$.
\end{definition}

\begin{definition}[Group Importance Score]
For a group $\gG_j\in\sG$ with group mask $\mM^{\gG_j}\in\sM$, the group importance score vector is defined by the Fisher information as $\text{GI}(\gG_j)= \left(\nabla \log \sum_{\gT_i \in \gG_i}\gL_{\gT_i}\left( \theta\odot\mM^{\gG_j}\right) \odot \mM^{\gG_j} \right)^2$. The absolute value of parameters is not used to measure importance here, as it is less effective than Fisher information in HarmoDT (Table~\ref{tb:50task}).
\end{definition}

Similarly, the Group Harmony Score can be defined as:
\begin{equation} \nonumber
    \label{eq:GH}
    \!\!\text{GH}(\gG_j,\mM^{\gG_j})_k=\left\{
    \begin{array}{ll}
    \!\!\text{GA}(\gG_j)_k+\lambda \text{GI}(\gG_j)_k, & (\mM^{\gG_j})_k =1, \\
    \!\!\inf ,  & (\mM^{\gG_j})_k =0.\\
    \end{array} \right. 
\end{equation}

\subsubsection{Group-wise Mask Update} \label{sec:group_find}
As total framework depicted in Figure~\ref{fig:framework} and Algorithms~\ref{alg:groupdt}--\ref{alg:update}, cohesive tasks can be clustered into groups via a warming-up and grouping mechanism.
We initially warm up the weights for $t_w$ iterations, followed by computing the harmony score for all tasks:
\begin{equation}
\mathcal{H}=[\text{GH}(\gT_i,\mathbf{1}),\dots,\text{GH}(\gT_N,\mathbf{1})],
\end{equation}
where $N$ is the total task number, and $\text{GH}(\gT_i,\mathbf{1})$ represents the Group Harmony Score for a single task group~($|\sG|=|\sT|$) with an all-one vector mask $\mathbf{1}$.
With harmony scores, we can cluster tasks into $N^G$ groups using any clustering techniques:
\begin{equation}
\sG=\{\gG_1,\dots,\gG_{N^G}\}=\text{Clustering}(\mathcal{H},N^G).
\end{equation}
In the experiments, we adopt the $k$-Nearest-Neighbors~\citep{knn1,knn2} for clustering. Notably, our aim is not to propose a new clustering method, and the visualization provided in Section~\ref{sec:further} illustrates the effectiveness of the clustering. 
We then initialize all tasks in the same group with a group mask as
\begin{equation}
    \mM^{\gG_j}=\mathbf{1}-\operatorname{ArgBtmK}_{\mathrm{S}*|\theta|}\left(\sum_{\gT_i \in \gG_j} \text{GH}(\gT_i,\mathbf{1})\right),
\end{equation}
where $|\theta|$ is the total number of the parameters. This operation ensures that each mask adheres to the same sparsity requirements $\mathrm{S}$.

\begin{algorithm}[t!]
    {\footnotesize
    \caption{G-HarmoDT}
    \label{alg:groupdt}  
    \begin{algorithmic}
    \STATE {\bfseries Input:} A set of tasks $\sT = \{ \gT_1,\dots,\gT_N \}$, maximum iteration $E$, episode length $T$, target return $\tG^*$, learning rate $\eta$, hyper-parameters $\{\eta_{max}, \eta_{min}, t_m, \lambda, \mathrm{S}, t_w,  N^G, gn\}$.
    \STATE {\color{gray}\te{// grouping stage}} \\
    \STATE Initialize the parameters of the network $\theta$ and gating module $\theta_g$, $t = 1$.
    \WHILE{$t \leq t_w$}
       \STATE sample a task $\gT_i$ from the set of tasks $\sT$, and sample a batch of data $\tau_i$ from the dataset $\gD_i$ of $\gT_i$. 
        \STATE $\theta \leftarrow \theta - \eta \nabla \gL_{\gT_i}(\theta)$, $t \leftarrow t + 1$.
    \ENDWHILE
    \STATE \colorbox{red!20}{$\sM, \sG$=\textbf{Grouping}($\sT, \theta,\lambda, N^G, \mathrm{S}$).}
    \STATE {\color{gray}\te{// training stage}} \\
    \WHILE{$t <E$}
        \STATE $\alpha_t=\lceil\eta_{max}+\frac{1}{2}(\eta_{min}-\eta_{max})(1+\cos{(2\pi \frac{t}{E})})\rceil$.
        \STATE \colorbox{blue!20}{$\sM$=\textbf{Group-wise\_Mask\_Update}($\sT, \sG, \sM, \theta, \alpha_t, \lambda$).}
        \WHILE{$t$ mod $t_m \neq 0$}
            \STATE sample a task $\gT_i$~($\gT_i\in \gG_j$) from $\sT$, and sample a batch of data $\tau_i$ from the dataset $\gD_i$ of $\gT_i$. 
            \STATE $\theta \leftarrow \theta - \eta \nabla \gL_{\gT_i}(\theta \odot \mM^{\gG_j}) \odot \mM^{\gG_j}$, $t \leftarrow t + 1$.
        \ENDWHILE 
    \ENDWHILE
    \STATE Train the Gating network $\theta_g$ with Equation \ref{eq:gateinput} as input and $\sG$ as target with cross-entropy loss. \\
    \STATE {\color{gray}\te{// inference stage}} \\
    \FOR{$i = 1, \dots, N$}
        \STATE Initialize history $\tau$ with zeros, desired reward $\hat{r} = \tG_i^*$, prompt $\tau^*$. \\
        \STATE Sample $3 \times gn$ length of trajectories with parameters $\theta$ and input to the Gating network, output current group $\gG$. \\
        \STATE Assign the parameters $\theta_i \leftarrow \theta \odot \mM^{\gG}, t \leftarrow 1$.
        \FOR{$t \leq T$}
            \STATE Get action $a = \text{Transformer}_{\theta_i}(\tau^*, \tau)[-1]$.
            \STATE Step env. and get feedback $\rvs, \rva, r, \hat{r} \leftarrow \hat{r} - r$.
            \STATE Append $[\rvs, \rva, \hat{r}]$ to recent history $\tau$.
            \STATE $t \leftarrow t + 1$.
        \ENDFOR
    \ENDFOR
    \end{algorithmic}
 }
\end{algorithm}

Building on the original processes of HarmoDT defined in \eqref{eq:weightmask} and \eqref{eq:weightrecover}, and adapting them into group level, we iteratively evaluate weights, mask group subspaces as
\begin{equation}
\label{eq:grow}
    \mM^{\gG_j}=\mM^{\gG_j}-\operatorname{ArgBtmK}_{\alpha_t}\left(\text{GH}(\gG_j, \mM^{\gG_j})\right),
\end{equation}
and recover the same number of group subspace as
\begin{equation}\label{eq:dead}
\begin{aligned}
\mM^{\gG_j}=&\mM^{\gG_j}+\operatorname{ArgTopK}_{\alpha_t}(
\sum_{\gT_i \in \gG_j}\rvg_{i}(\theta)\odot \\
&\frac{1}{N^G} \sum_{\gG_k \in \sG} \sum_{\gT_i\in \gG_k}\bar{\rvg}_{i}(\theta)- \inf \odot \mM^{\gG_j}).
\end{aligned}
\end{equation}
We provide the framework and detailed procedures in the third stage of Figure~\ref{fig:framework} and Algorithm~\ref{alg:update}.

\subsubsection{Gating Module}
Since we lack access to the task identifier of the current task, we depend on a gating network to ascertain the current task. 
In the RL environment, formulated as the MDP, each task consists of multiple transitions, encompassing the state $\rvs$, action $\rva$, and reward $r$, which serve as the input to the gating network. 
To streamline the gating network, we formulate it as an MLP-based model comprising three layers with ReLU activations, utilizing 128 hidden units for all layers except the input layer.
The input to the gating network comprises concatenated states, actions, and rewards:
\begin{equation}
\label{eq:gateinput}
\text{Input} = (\rvs_{1} ~||~ \rva_{1} ~||~ r_1 ~||~ \dots ~||~ \rvs_{gn} ~||~ \rva_{gn} ~||~ r_{gn}),
\end{equation}
where $gn$ is the hyper-parameter, and the output is the group ID of the current task.


\subsubsection{Total Framework}
\label{sec:inferofg}
As illustrated in Figure~\ref{fig:framework}, G-HarmoDT initially warm up the weights, followed by clustering cohesive tasks into the same group based on the Group Harmony Score. Subsequently, all tasks are initialized with a group mask. With the group masks in place, the updating of model weights is defined as follows:
\begin{equation}
    \theta_{t+1}=\theta_{t}-\eta \E_{\gT_i \sim p(\gT)} \nabla \gL_{\gT_i\in\gG_j}(\theta \odot \mM^{\gG_j}) \odot \mM^{\gG_j},
\end{equation}
using a loss function similar to that in Equation \ref{eq:dtloss}.
At every group mask updating interval $t_m$, we update the group mask as defined in Equations \ref{eq:grow} and \ref{eq:dead}.
Simultaneously, we update the gating network using task inputs and the ID-Group Mapping $\sG$ as labels, optimizing with the cross-entropy loss.
During the inference stage without task identifiers, we initially acquire the group identifier through a few transitions obtained from interactions with the simulator. 
Based on the gating network's results, we load the corresponding parameter subspace for the group and use it to iteratively make decisions based on the current state in an auto-regressive manner.

\begin{algorithm}[t!]
    {\footnotesize
    \caption{Grouping}
    \label{alg:Grouping}
    \begin{algorithmic}
    \STATE {\bfseries Input:} A set of tasks $\sT$, trainable weights vector $\theta$, and hyper-parameters $\{ \lambda, N^G, \mathrm{S}\}$.
    \FOR{$i= 1,\dots,N$}
        \STATE $\rvg_i= \nabla \gL_{T_i} (\theta) $.
        \STATE $\bar{\rvg}_i= \nabla \gL_{\gT_i} (\theta\odot \mM^{\gG_j})\odot \mM^{\gG_j}, ~ \gT_i \in \gG_j $.
    \ENDFOR
    \FOR{$i= 1,\dots,N$}
        \STATE Calculate $\text{GH}(\gT_i, \mathbf{1})$ with $\lambda$, $\rvg_i$ and $\bar{\rvg}$ as Sec.~\ref{sec:groupcalculate}.
    \ENDFOR
    \STATE $\mathcal{H}=[\text{GH}(\gT_i,\mathbf{1}),\dots,\text{GH}(\gT_n,\mathbf{1})]$.
    \STATE $\sG=\{\gG_1,\dots,\gG_{N^G}\}=\text{Clustering}(\mathcal{H},N^G)$.
    \FOR{$i= 1,\dots,N^G$}
        \STATE $\mM^{\gG_j}=\mathbf{1}-\operatorname{ArgBtmK}_{\mathrm{S}*|\theta|}\left(\sum_{\gT_i \in \gG_j} \text{GH}(\gT_i,\mathbf{1})\right)$.
    \ENDFOR
    \STATE $\sM=\{\mM^{\gG_1},\dots,\mM^{\gG_{N^G}}\}$.
    \end{algorithmic}
{\textbf{Output}:$\sM,\sG.$}
}
\end{algorithm}

\begin{algorithm}[t!]
    {\footnotesize
    \caption{Group-wise Mask Update}
    \label{alg:update}
    \textbf{Input}:A set of tasks $\sT$, groups $\sG$, group masks $\sM$, trainable weights $\theta$, updating number $\alpha_t$, and hyper-parameters $\{\lambda\}$.
    \begin{algorithmic}
    \FOR{$i= 1,\dots,N$}
        \STATE $\rvg_i= \nabla \gL_{\gT_i} (\theta) $.
        \STATE $\bar{\rvg}_i= \nabla \gL_{\gT_i} (\theta\odot \mM^{\gG_j})\odot \mM^{\gG_j}, ~ \gT_i \in \gG_j $.
    \ENDFOR
    \FOR{$j= 1,\dots, N^G$}
        \STATE Calculate $\text{GH}(\gG_j, \mM^{\gG_j})$ with $\lambda$, $\rvg_i$ and $\bar{\rvg}$ as Sec.~\ref{sec:groupcalculate}.
        \STATE $\mM^{\gG_j}=\mM^{\gG_j}-\operatorname{ArgBtmK}_{\alpha_t}\left(\text{GH}(\gG_j, \mM^{\gG_j})\right)$.
        \STATE $\mM^{\gG_j}=\mM^{\gG_j}+\operatorname{ArgTopK}_{\alpha_t}(\sum_{\gT_i \in \gG_j}\rvg_{i}(\theta)\odot \frac{1}{N^G} \sum_{\gG_k \in \sG} \sum_{\gT_i\in \gG_k}\bar{\rvg}_{i}(\theta)- \inf \odot \mM^{\gG_j})$.
    \ENDFOR
    \end{algorithmic}
    {\textbf{Output}:$\sM = \{\mM^{\gG_1},\dots,\mM^{\gG_{N^G}}\}$.}
 }
\end{algorithm}
\section{Experiment}\label{sec:exp}
In this section, we conduct extensive experiments to answer the following questions\footnote{Our code is available at: \url{https://github.com/charleshsc/HarmoDT}}:
(1) How does our method compare to other offline and online baselines in the multi-task regime?
(2) Does our method mitigate the phenomenon of conflicting gradients and identify optimal harmony subspaces of parameters for tasks?
(3) Can our method generalize to unseen tasks and G-HarmoDT effectively reduce the need for task identifier?

\subsection{Experimental Setups}\label{sec:env}

\subsubsection{Environment}
Our experiments utilize the Meta-World benchmark \citep{yu2020meta}, featuring 50 distinct manipulation tasks with shared dynamics, requiring a Sawyer robot to interact with various objects. We extend tasks to a random-goal setting, consistent with recent studies \citep{he2023not, sun2022paco}.
Performance evaluation is based on the averaged success rate across tasks. 
Following \citet{he2023diffusion}, we employ two dataset compositions: a near-optimal dataset from SAC-Replay \citep{sac} ranging from random to expert experiences, and a sub-optimal dataset with initial trajectories and a reduced proportion (50\%) of expert data.

For unseen tasks, we primarily evaluate HarmoDT on distinct objectives drawn from datasets used in previous works \citep{mitchell2021offline, yu2020meta, PDT, PTDT}, specifically Cheetah-dir, Cheetah-vel, and Ant-dir, which challenge the agent to optimize direction and speed. 
Details on environment specifications and hyper-parameters are available in Appendix~B.

\subsubsection{Baselines}\label{sec:expbaseline}
We compare our methods against the following baselines:
(i) \textbf{MTBC}: Extends Behavior Cloning for multi-task learning with network scaling and a task-ID conditioned actor;
(ii) \textbf{MTIQL}: Adapts IQL \citep{IQL} with multi-head critics and a task-ID conditioned actor for multi-task policy learning;
(iii) \textbf{MTDIFF-P} \citep{he2023diffusion}: A diffusion-based method combining Transformer architectures and prompt learning for generative planning in multitask offline settings;
(iv) \textbf{MTDT}: Extends DT \citep{DT} to multitask settings, utilizing a task ID encoding and state input for task-specific learning;
\textbf{MGDT} \citep{lee2022multi}: An extension of DT \citep{DT} designed to regulate action generation, consistently producing high-reward behaviors based on Bayesian principles;
(vi)\textbf{Prompt-DT} \citep{PDT}: Builds on DT, leveraging trajectory prompts and reward-to-go for multi-task learning and generalization to unseen tasks.
(vii) \textbf{MTDT+PCGrad \citep{yu2020gradient} \& CAGrad \citep{liu2021conflict}}: Augments MTDT with gradient surgery techniques to resolve gradient conflicts.
We also compare our method with four online RL approaches:
(viii) \textbf{CARE} \citep{sodhani2021multi}: Utilizes metadata and a mixture of encoders for task representation;
(ix) \textbf{PaCo} \citep{sun2022paco}: Employs a parameter compositional approach for task-specific parameter recombination;
(x) \textbf{Soft-M} \citep{yang2020multi}: Focuses on a routing network for the soft combination of modules;
(xi) \textbf{D2R} \citep{he2023not}: Adopts disparate routing paths for module selection per task.
Unless otherwise specified, HarmoDT and G-HarmoDT use Fisher information to calculate weight importance.

\begin{table}[!t]
\setlength{\tabcolsep}{3pt}
\centering
\caption{Average success rate across 3 seeds on Meta-World MT50 with random goals  under both near-optimal and sub-optimal cases.
Each task is evaluated for 50 episodes.
Approaches with * indicate baselines of our own implementation.
}
\vspace{0.1cm}
\small
\centering
\scalebox{1.00}{
\begin{tabular}{l|ccc}
\toprule[2pt]
Method& \multicolumn{3}{c}{ Meta-World 50 Tasks} \\
\cmidrule{1-4} \#Partition &Near-optimal  & Sub-optimal & Params\\
\midrule 
CARE~(online) & $46.12_{\pm 1.30}$ & \textbf{-}&1.26 M \\
PaCo~(online) & $54.31_{\pm 1.32}$ & \textbf{-}&3.39 M\\
Soft-M~(online) & $53.41_{\pm 0.72}$ & \textbf{-}& 1.62 M\\
D2R~(online) & $63.53_{\pm 1.22}$ & \textbf{-}& 1.40 M\\
\midrule
PCGrad*  & $46.34_{\pm 1.73}$ & $27.04_{\pm 2.11}$&1.74 M\\
CAGrad*  & $45.60_{\pm 1.34}$ & $27.63_{\pm 1.82}$&1.74 M\\
MTBC  & $60.39_{\pm 0.86}$ & $34.53_{\pm 1.25}$&1.74 M\\
MTIQL & $56.21_{\pm 1.39}$ & $43.28_{\pm 0.90}$&1.74 M \\
MTDIFF-P & $59.53_{\pm 1.12}$ & $48.67_{\pm 1.32}$&5.32 M \\
MTDIFF-P-ONEHOT& $61.32_{\pm 0.89}$ & $\textbf{48.94}_{\pm 0.95}$&5.32 M \\
MTDT & $20.99_{\pm 2.66}$ & $20.63_{\pm 2.21}$& 0.87 M\\
MTDT* & $ 65.80_{\pm 1.02}$ & $42.33_{\pm 1.89}$& 1.47 M \\
MGDT*  & $66.63_{\pm 1.11}$ & $42.82_{\pm 1.03}$&2.15 M\\
Prompt-DT & $45.68_{\pm 1.84}$ & $39.76_{\pm 2.79}$ &0.87 M\\
Prompt-DT* & $\textbf{69.33}_{\pm 0.89}$ & $48.40_{\pm 0.16}$&1.47 M \\
\midrule
HarmoDT-R~(ours) & $\textbf{75.39}_{\pm 1.18}$ & $\textbf{53.80}_{\pm 1.07}$&1.47 M \\
HarmoDT-M (ours) & $\textbf{80.33}_{\pm 0.97}$ & $\textbf{57.20}_{\pm 0.73}$ &1.47 M\\
HarmoDT-F (ours) & $\textbf{82.20}_{\pm 0.40}$ & $\textbf{57.20}_{\pm 0.68}$ &1.47 M\\
\bottomrule[2pt]
\end{tabular}
}
\label{tb:50task}
\vspace{-0.5cm}
\end{table}

\begin{table*}[t!]
\centering
\caption{Average success rate of 50 episodes across 3 seeds under task provided and agnostic cases. We randomly select 5, 30, and 50 tasks from Meta-World under near-optimal and sub-optimal cases. "/" indicates items that are not applicable to the method.}
\label{tb:main}
\small
\centering
\scalebox{1.0}{
\begin{tabular}{l|l|cccccc}
\toprule[2pt]
\multirow{2}{*}[-4pt]{ Env. Type }& \#Tasks & \multicolumn{2}{c}{ MT5 } & \multicolumn{2}{c}{MT30} & \multicolumn{2}{c}{MT50}\\
\cmidrule{2-8} & \#Settings& near-optimal & sub-optimal& near-optimal & sub-optimal& near-optimal & sub-optimal\\
\midrule 
\multirow{7}{*}[0pt]{Task-provided} 
&MTDT & \textbf{100.0$_{\pm 0.0}$} & 64.7$_{\pm 5.3}$ & 71.9$_{\pm 1.0}$ & 49.3$_{\pm 2.1}$ & 65.8$_{\pm 1.0}$ & 42.3$_{\pm 1.9}$ \\
& MGDT & \textbf{100.0$_{\pm 0.0}$} & 66.7$_{\pm 2.2}$& 71.3$_{\pm 1.9}$ & 49.3$_{\pm 1.4}$& 66.6$_{\pm 1.1}$ & 42.8$_{\pm 1.0}$ \\
&PCGrad & \textbf{100.0$_{\pm 0.0}$} & 74.2$_{\pm 1.7}$ & $62.3_{\pm 1.4}$ & $40.2_{\pm 2.0}$ & $46.3_{\pm 1.7}$ & $27.0_{\pm 2.1}$  \\
&CAGrad & \textbf{100.0$_{\pm 0.0}$} & $72.3_{\pm 2.1}$ & $66.4_{\pm 1.2}$ & $42.4_{\pm 1.9}$ & $45.6_{\pm 1.3} $ & $27.6_{\pm 1.8}$ \\
&PromptDT & \textbf{100.0$_{\pm 0.0}$} & 66.5$_{\pm 2.7}$ & 74.1$_{\pm 0.7}$ & 53.8$_{\pm 0.6}$& 69.3$_{\pm 0.9}$ & 48.4$_{\pm 0.2}$ \\
 &MTDIFF & \textbf{100.0$_{\pm 0.0}$} & 66.3$_{\pm 2.3}$& 67.5$_{\pm 0.4}$ & 54.2$_{\pm 1.1}$ & 61.3$_{\pm 0.9}$ & 48.9$_{\pm 1.0}$ \\
  \cmidrule{2-8}
 & HarmoDT & \textbf{100.0$_{\pm 0.0}$} & \textbf{77.7$_{\pm 3.3}$} & \textbf{88.0$_{\pm 1.0}$} & \textbf{65.0$_{\pm 0.9}$} & \textbf{82.2$_{\pm 0.4}$} & \textbf{57.2$_{\pm 0.7}$} \\
\midrule 
\multirow{12}{*}{Task-agnostic} 
&MTDT & \textbf{100.0$_{\pm 0.0}$} & 62.0$_{\pm 2.4}$& 69.3$_{\pm 1.7}$ & 46.3$_{\pm 2.0}$& 60.8$_{\pm 1.2}$ & 39.8$_{\pm 2.1}$ \\
& MGDT & \textbf{100.0$_{\pm 0.0}$} & 64.4$_{\pm 3.1}$& 69.7$_{\pm 2.1}$ & 48.3$_{\pm 2.4}$& 62.0$_{\pm 1.9}$ & 38.2$_{\pm 2.2}$ \\
& PCGrad & \textbf{100.0$_{\pm 0.0}$} & 72.0$_{\pm 0.7}$& 52.3$_{\pm 0.8}$ & 37.7$_{\pm 1.0}$& 44.8$_{\pm 0.6}$ & 25.8$_{\pm 1.3}$ \\
& CAGrad & \textbf{100.0$_{\pm 0.0}$} & 71.2$_{\pm 0.5}$ & $51.7_{\pm 0.6}$ & $39.2_{\pm 1.3}$ & $41.4_{\pm 0.9}$ & $26.2_{\pm 1.1}$\\

& PromptDT & 84.3$_{\pm 0.9}$ & 48.0$_{\pm 2.3}$ & 44.7$_{\pm 1.7}$ & 35.7$_{\pm 1.4}$& 34.2$_{\pm 2.2}$ & 22.2$_{\pm 2.9}$ \\
 & MTDIFF & 76.0$_{\pm 1.7}$ & 62.0$_{\pm 1.4}$ & 42.3$_{\pm 1.1}$ & 41.0$_{\pm 0.9}$ & 38.0$_{\pm 1.5}$ & 23.2$_{\pm 1.2}$ \\
    \cmidrule{2-8}
  & HarmoDT& 46.0$_{\pm 2.1}$ & 30.0$_{\pm 1.9}$& 17.0$_{\pm 1.2}$ & 15.7$_{\pm 1.6}$& 16.8$_{\pm 1.4}$ & 15.4$_{\pm 1.9}$ \\
  & G-HarmoDT-5 & \textbf{100.0$_{\pm 0.0}$} & \textbf{77.7$_{\pm 3.3}$} & \textbf{73.7$_{\pm 2.7}$} & 51.7$_{\pm 2.3}$ & 59.8$_{\pm 1.8}$ & 40.4$_{\pm 1.4}$\\
  & G-HarmoDT-10 & / & / & 71.0$_{\pm 1.8}$ & \textbf{54.7$_{\pm 1.5}$} & \textbf{69.6$_{\pm 1.7}$} & 42.0$_{\pm 1.3}$\\
 & G-HarmoDT-30 & / & / & 69.3$_{\pm 1.4}$ & 47.0$_{\pm 2.4}$ & 65.0$_{\pm 2.1}$ & \textbf{45.8$_{\pm 1.6}$} \\
 & G-HarmoDT-50 & / & / & / & / & 62.6$_{\pm 1.9}$ & 37.2$_{\pm 1.1}$\\
\bottomrule[2pt]
\end{tabular}
}
\end{table*}

\subsection{Main Results}
This section presents the superior performance of HarmoDT in task-provided and unseen settings, and of G-HarmoDT in task-agnostic settings, compared to extensive baselines, along with an analysis of scalability across task scales.

\subsubsection{Multi-task Performance}
For multi-task performance, we benchmark HarmoDT against all established baselines introduced in Section~\ref{sec:expbaseline} on 50 Meta-World tasks. We consider various weight evaluation methods introduced in Section~\ref{sec:harmo_metric}, including \textbf{HarmoDT-R}, which keeps task masks frozen; \textbf{HarmoDT-F}, which uses Fisher information $I_F(\gT_i)$ to calculate weight importance; and \textbf{HarmoDT-M}, which relies on magnitude information $I_M(\gT_i)$ for determining weight importance. 

As shown in Table~\ref{tb:50task}, with fixed random masks, HarmoDT-R surpasses all other methods, achieving a 6.1\% and 4.9\% improvement in near-optimal and sub-optimal scenarios, respectively, compared to the best baseline. 
Furthermore, HarmoDT-M and HarmoDT-F enhance HarmoDT-R's performance by learning task masks to identify optimal harmony subspaces, resulting in substantial gains of 6.8\% and 3.4\% in near-optimal and sub-optimal cases, respectively.
This demonstrates the effectiveness of HarmoDT in multi-task settings, handling both sub-optimal datasets that require stitching useful segments from suboptimal trajectories, and near-optimal datasets where emulating optimal behaviors is essential.

\subsubsection{Task-agnostic Performance}
In task-agnostic settings, we show results of G-HarmoDT compared to baselines designed for multi-task learning, including MTDT, MGDT, PCGrad, CAGrad, PromptDT, and MTDIFF. 
As depicted in Table~\ref{tb:main}, most methods struggle to achieve satisfactory results without task IDs, with HarmoDT notably experiencing an average drop of 54.6\% in task success rates. 
However, through the integration of grouping and gating modules, our G-HarmoDT consistently outperforms all other methods across a broad spectrum of group numbers, demonstrating significant superiority: 5.7\% for 5 tasks, 6.4\% for 30 tasks, and 6\% for 50 tasks in sub-optimal settings. 
It is noteworthy that performance does not always improve with an increasing number of groups, as it becomes more challenging for the gating module to accurately identify more groups, resulting in a degradation in overall performance.
Together with the ablation results shown in Table~\ref{tb:gating} and Figure \ref{fig:reduce_burden}, this illustrates that simply combining the gating module with existing ID-dependent methods does not yield the best performance.
However, when coupled with a grouping mechanism, our G-HarmoDT demonstrates superior effectiveness in task-agnostic settings.

\subsubsection{Scalability of Task Scale.} 
We also evaluated the scalability of HarmoDT and G-HarmoDT across varying task numbers under both task-provided and task-agnostic settings in the Meta-World benchmark. 
As shown in Table~\ref{tb:main} and Figure \ref{fig:performance_num}, HarmoDT consistently outperforms MTDIFF, MTDT, and Prompt-DT across all task numbers under task-provided settings, with notable improvements as the task count increases: 11\% for 5 tasks, 11\% for 30 tasks, and 8\% for 50 tasks in sub-optimal cases. 
In task-agnostic settings, G-HarmoDT consistently surpasses existing methods, showing an average improvement of 9.4\% over the best baseline across all scales.

\begin{table}[t!]
\setlength{\tabcolsep}{4pt}
\centering
\caption{Comparison of generalization ability. We conduct experiments and record cumulative reward of unseen tasks on three distinct datasets: Cheetah-dir, Cheetah-vel, and Ant-dir, which challenge the agent to optimize direction and speed.}
\vspace{2pt}
\label{tb:unseen}
\small
\centering
\scalebox{1.0}{
\begin{tabular}{l|ccc}
\toprule[2pt]
Setting & MTDT & Prompt-DT & HarmoDT-F \\
\midrule 
Cheetah-dir & 662.40$_{\pm 1.3}$ & 935.3$_{\pm 2.6}$& \textbf{958.5}$_{\pm 1.5}$ \\
Cheetah-vel & -170.11$_{\pm 5.7}$ & -127.7$_{\pm 9.9}$& \textbf{-66.51}$_{\pm 1.2}$\\
Ant-dir & 165.29$_{\pm 0.4}$ & 278.7$_{\pm 38.7}$& \textbf{298.3}$_{\pm 1.0}$ \\
\midrule 
Average & 219.2 & 362.1& \textbf{396.8}$_{9.6\% \uparrow}$ \\
\bottomrule[2pt]
\end{tabular}
}
\vspace{-0.4cm}
\end{table}

\subsubsection{Generalization Performance on Unseen tasks}
Prompt-DT's proficiency with unseen tasks prompted us to assess HarmoDT's capabilities in similar scenarios. 
We employ a voting mechanism among all observed tasks to define a generalized subspace for unseen tasks, as delineated in Equation \ref{eq:unseen}. 
This technique operates on a foundational assumption: parameters that are consistently identified as significant and harmonious across a range of tasks (surpassing a predefined threshold) are posited to hold universal value, potentially contributing positively to task performance in novel scenarios. 
Comparative analysis involving HarmoDT, MTDT, and Prompt-DT is conducted on three distinct datasets: Cheetah-dir, Cheetah-vel, and Ant-dir.
The results, presented in Table~\ref{tb:unseen}, affirm HarmoDT's comprehensive enhancements across all test cases. 
Notably, HarmoDT achieves an average reward of 396.8, a 9.6\% increase over Prompt-DT's 362.1, highlighting the effectiveness of our voting approach in handling unseen tasks.

\subsection{Further Analysis}
\label{sec:further}
This section explores ablation studies on gating module, hyper-parameters, model size, computational times, and visualizations of learned masks and the grouping results for tasks produced by HarmoDT and G-HarmoDT, respectively.
\begin{table}[t!]
\centering
\caption{
Ablation study of gating module. We apply the gating module to ID-dependent approaches and compare their performance on Meta-World 50 tasks under task-agnostic settings.
}
\small
\centering
\scalebox{1.00}{
\begin{tabular}{l|cc}
\toprule[2pt]
Method& \multicolumn{2}{c}{ Meta-World 50 Tasks} \\
\cmidrule{1-3} \#Partition &Near-optimal  & Sub-optimal\\
\midrule 
PromptDT  & 34.2$_{\pm 2.2}$ & 22.2$_{\pm 2.9}$\\
\textbf{+gating}  & 49.2$_{\pm 1.7}$ & 34.8$_{\pm 1.9}$\\
MTDIFF & 38.0$_{\pm 1.5}$ & 23.2$_{\pm 1.2}$ \\
\textbf{+gating}  & 44.1$_{\pm 2.1}$ & 35.2$_{\pm 3.2}$\\
\midrule 
HarmoDT (Ours)& 16.8$_{\pm 1.4}$ & 15.4$_{\pm 1.9}$ \\
\textbf{+gating}  & 62.6$_{\pm 1.9}$ & 37.2$_{\pm 1.1}$\\
G-HarmoDT-10 (Ours) & \textbf{69.6}$_{\pm 1.7}$ & \textbf{45.8}$_{\pm 1.6}$\\
\bottomrule[2pt]
\end{tabular}
}
\label{tb:gating}
\end{table}

\begin{table}[t!]
\setlength{\tabcolsep}{4pt}
\centering
\caption{Ablation study on the model size of MTDT, Prompt-DT, and our HarmoDT-F under near-optimal of Meta-World 30 tasks and 50 tasks. We denote the model with z M parameters and x layers of y head attentions as (x, y, z) in the table.}
\vspace{2pt}
\label{tab:modelsize}
\small
\centering
\scalebox{1.0}{
\begin{tabular}{c|c|ccc}
\toprule[2pt]
 Case & Model & MTDT & Prompt-DT & HarmoDT-F \\
\midrule 
\multirow{4}{*}[0pt]{30 tasks} 
 & (1, 2, 0.48) & 35.78$_{\pm 0.9}$ & 42.10$_{\pm 3.1}$& 59.33$_{\pm 1.5}$\\
 & (3, 4, 0.87) & 64.01$_{\pm 1.4}$ & 69.56$_{\pm 1.0}$ &77.67$_{\pm 1.6}$\\
 & (6, 8, 1.47) & 71.89$_{\pm 1.0}$ & 74.10$_{\pm 0.7}$& 88.00$_{\pm 1.0}$ \\
 & (9, 16, 2.06) & 74.33$_{\pm 0.7}$ & 77.67$_{\pm 1.7}$& 88.00$_{\pm 1.6}$ \\
\midrule
  \multirow{4}{*}[0pt]{50 tasks} 
 & (1, 2, 0.48) & 31.93$_{\pm 3.4}$ & 36.07$_{\pm 2.5}$& 58.20$_{\pm 1.9}$\\
 & (3, 4, 0.87) & 58.13$_{\pm 3.6}$ & 61.93$_{\pm 0.4}$ &72.80$_{\pm 1.6}$\\
 & (6, 8, 1.47) & 65.80$_{\pm 1.0}$ & 69.33$_{\pm 0.9}$& 82.20$_{\pm 0.4}$ \\
 & (9, 16, 2.06) &68.33$_{\pm 1.1}$ & 71.00$_{\pm 0.9}$& 83.33$_{\pm 0.6}$ \\
\bottomrule[2pt]
\end{tabular}
}
\vspace{-0.4cm}
\end{table}

\subsubsection{Ablation of gating}
Since G-HarmoDT incorporates a gating module to assign group IDs, in Table~\ref{tb:gating}, we present ablation results for the gating module applied to ID-dependent methods across 50 tasks in the Meta-World benchmark under task-agnostic settings.
The results demonstrate that our gating module significantly enhances task ID-dependent approaches. 
Furthermore, G-HarmoDT, which incorporates both grouping and gating modules, further enhances performance.
This improvement stems from G-HarmoDT's ability to reduce the gating module's learning burden and enhance identifier accuracy by converting task identification into group identification as shown in Figure~\ref{fig:reduce_burden}.

\begin{figure*}[t!]
    \centering
    \subfigure{
    \centering
    \label{fig:s3}
    \includegraphics[width=0.23\textwidth]{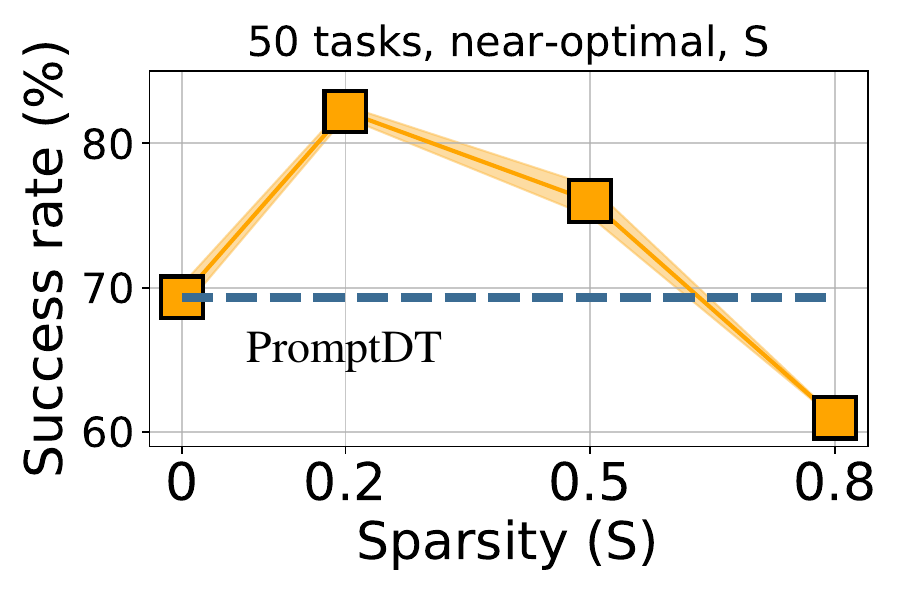}}
    \centering
    \subfigure{
    \centering
    \label{fig:e3}
    \includegraphics[width=0.23\textwidth]{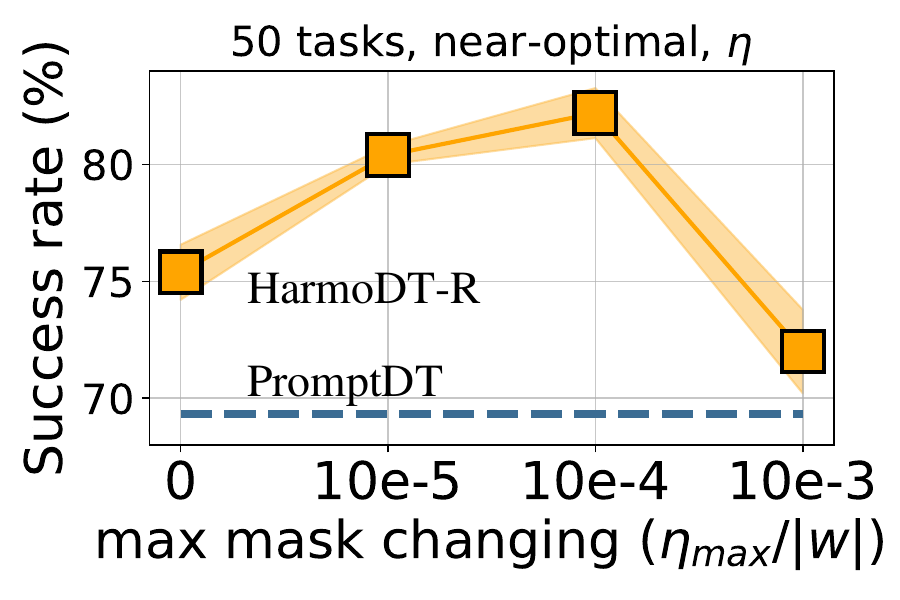}}
    \subfigure{
    \centering
    \label{fig:l3}
    \includegraphics[width=0.23\textwidth]{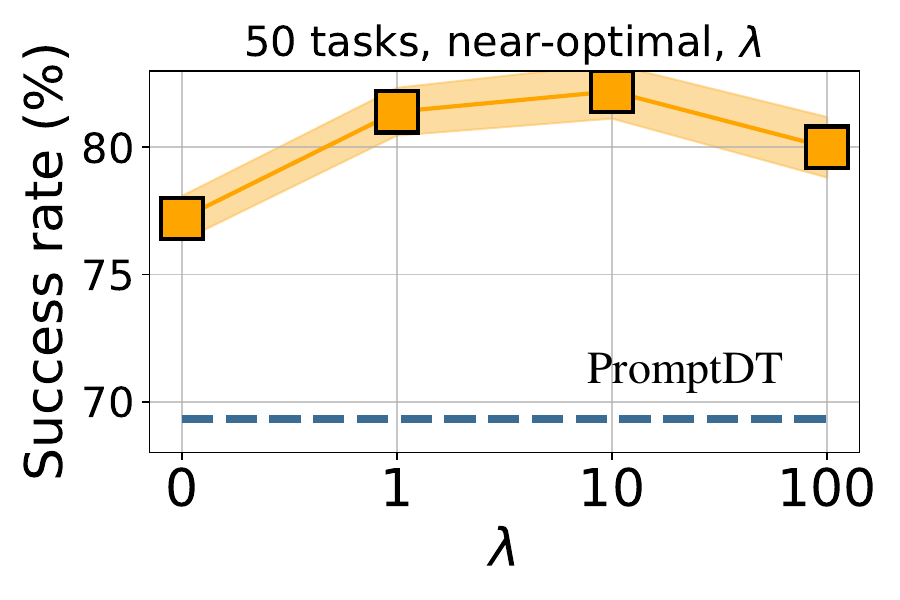}}
    \vspace{-0.4cm}
    \subfigure{
    \centering
    \label{fig:t3}
    \includegraphics[width=0.23\textwidth]{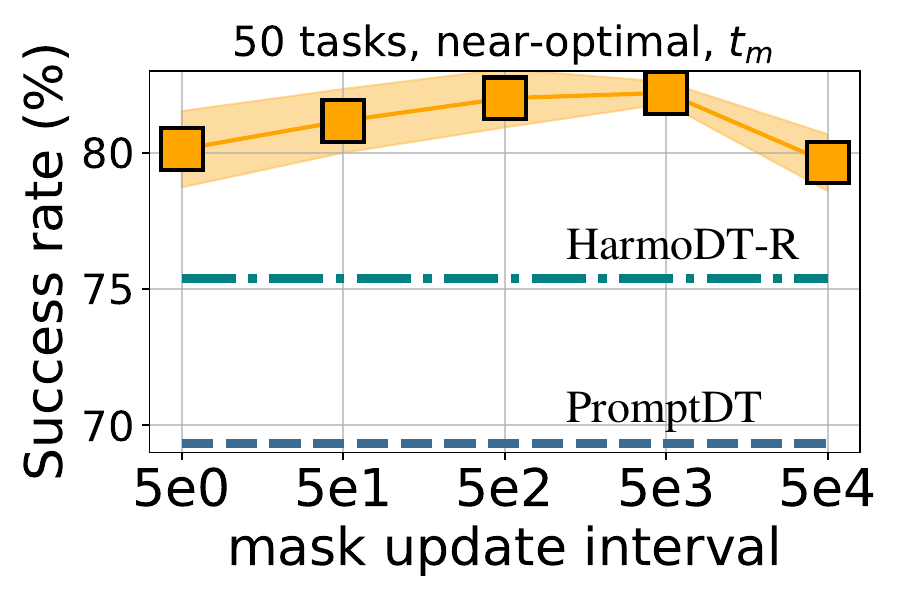}}    
     \subfigure{
    \centering
    \label{fig:g3}
    \includegraphics[width=0.23\textwidth]{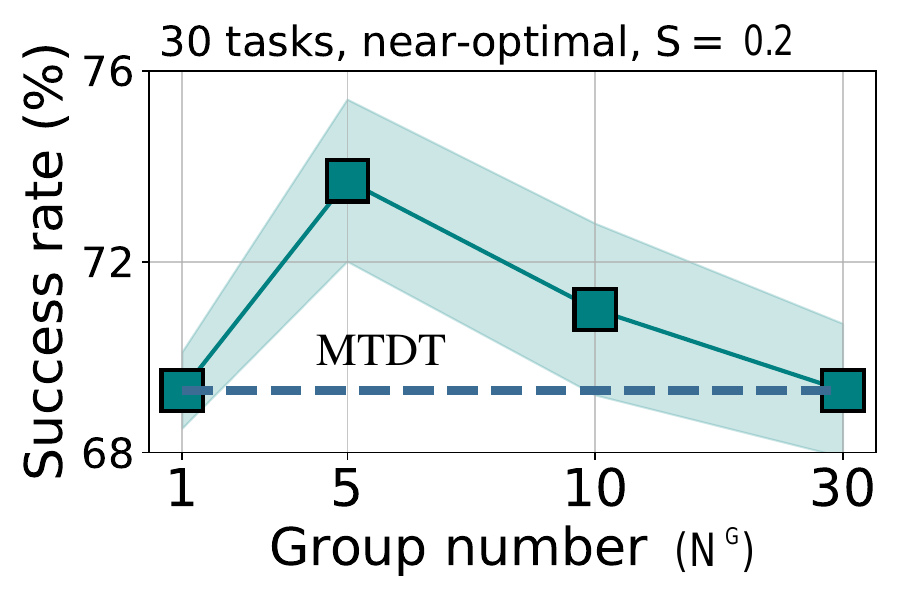}}
     \subfigure{
    \centering
    \label{fig:w3}
    \includegraphics[width=0.23\textwidth]{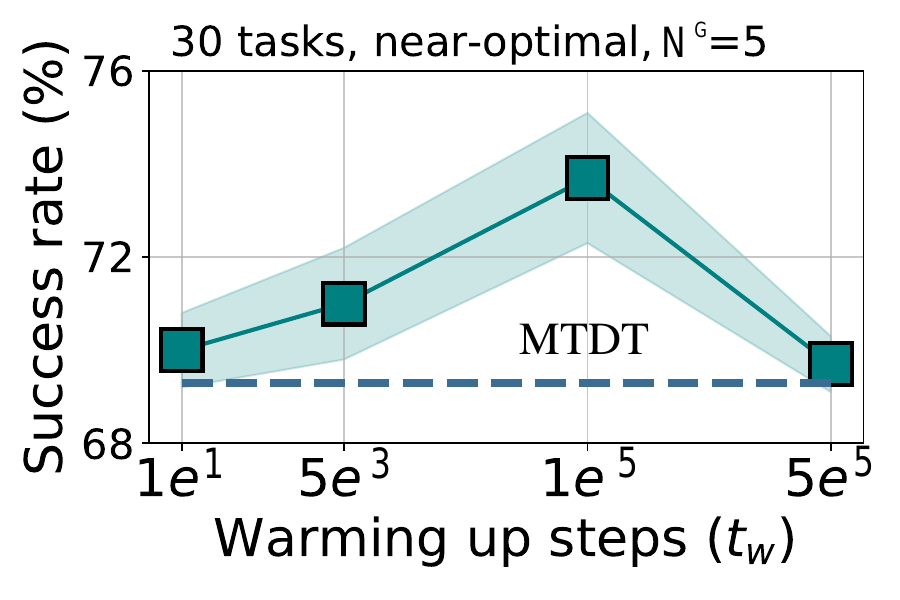}}
    \subfigure{
    \centering
    \label{fig:gg3}
    \includegraphics[width=0.23\textwidth]{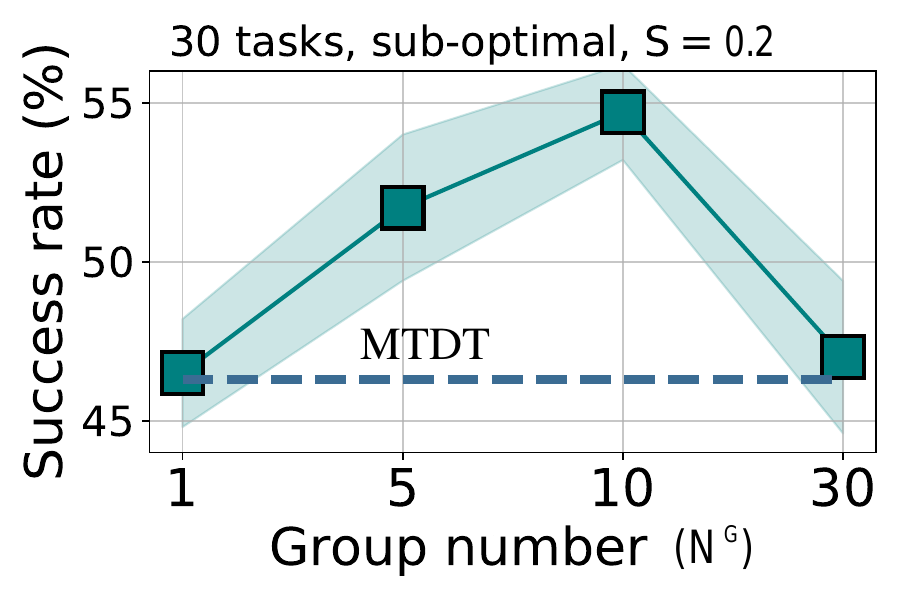}}
     \subfigure{
    \centering
    \label{fig:ww3}
    \includegraphics[width=0.23\textwidth]{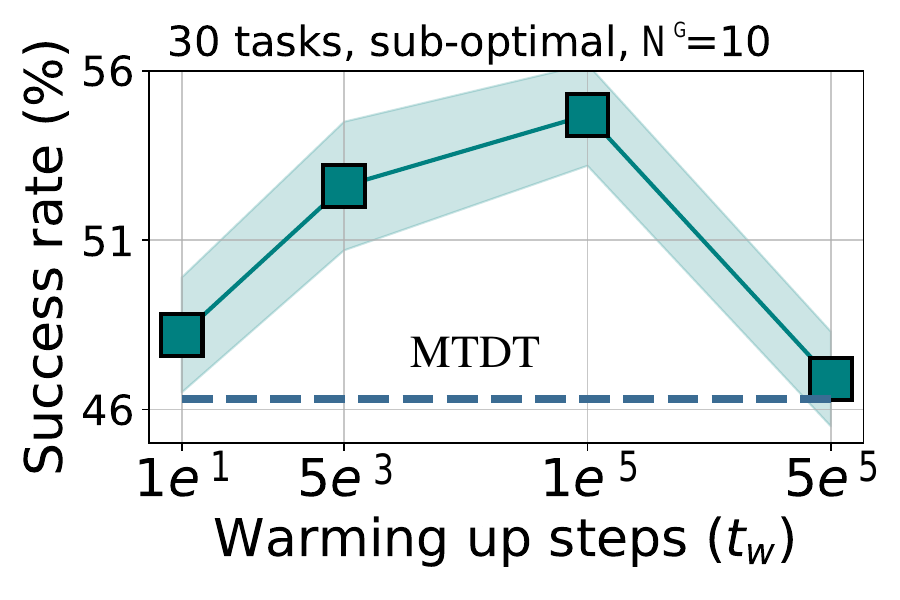}}
    \vspace{-0.2cm}
    \caption{The first row illustrates HarmoDT's ablation results of mask sparsity $\mathrm{S}$, maximum of mask changing within mask update interval $\eta_{max}$, balance controller $\lambda$, and mask update interval $t_m$ on the Meta-World benchmark with 50 tasks under the near-optimal case. 
    The second row illustrates G-HarmoDT's ablation results of the number of groups $N^G$, and the warming-up steps $t_w$ on the Meta-World benchmark with 30 tasks under task-agnostic settings and both the near-optimal and sub-optimal cases. 
    Default values are listed as $\eta_{min}=0$, $\eta_{max}$ is 100~(about 1e-3\% of total weights), $\mathrm{S}=0.2$, $\lambda=10$, $t_m=5e3$, $N^G=5$ and $t_w=1e5$.
    During each individual ablation, a single parameter is varied, with all other parameters maintained at their default values.}
    \label{fig:ablationofpara}
        \vspace{-0.3cm}
\end{figure*}
\subsubsection{Ablation on Hyper-parameters} 
This study introduces comprehensive ablations of selecting proper hyper-parameters for our method. 
The first row of Figure~\ref{fig:ablationofpara} delineates the ablation results of HarmoDT, including cosine annealing parameter $\eta_{max}$ (with $\eta_{min}=0$), mask alteration frequency $t_m$, overall sparsity $\mathrm{S}$, and balance controller $\lambda$ on 50 tasks of Meta-World benchmark in the near-optimal settings. 
The second row of Figure~\ref{fig:ablationofpara} illustrates the results of G-HarmoDT, including the number of groups $N^G$, and warming-up steps $t_w$ under task-agnostic settings and both the near-optimal and sub-optimal cases.
Based on these insights, recommended settings of hyper-parameters for HarmoDT are sparsity ratio $\mathrm{S}=0.2$, $\eta_{max}$ at approximately 0.001\% of total weights, balance factor $\lambda=10$ and mask changing interval $t_m=5000$ rounds. 
Additionally, for G-HarmoDT, recommended settings are approximately $1e5$ steps for $t_w$, and 5 or 10 for the number of groups $N^G$. 
Across a broad range of hyper-parameter values, our approach consistently outperforms baselines, with these parameters collectively contributing to its superior performance.

\begin{table}[t!]
\centering
\caption{Trainable parameter size and wall clock time on 50 tasks under near-optimal settings with one NVIDIA GeForce 4090.
}
\small
\centering
\scalebox{1.00}{
\begin{tabular}{l|ccc}
\toprule[2pt]
Method & Parameter Size & Wall-clock Time \\
\midrule 
MTDT & 1.47 M & 0d 21h 40m\\
PromptDT & 1.47 M  & 1d 02h 36m\\
MTDIFF & 5.32 M & 3d 01h 13m \\
MGDT & 2.15 M & 0d 21h 47m \\
\midrule
HarmoDT (Ours)& 1.47 M& 1d 02h 59m \\
G-HarmoDT-10 (Ours) & 1.87 M & 1d 03h 49m\\
\bottomrule[2pt]
\end{tabular}}
\label{tab:parasizetime}
\end{table}
\subsubsection{Model size and computational time}
The influence of model size is pronounced in multi-task training scenarios. 
Table \ref{tab:modelsize} delineates our ablation study on model size across 1e6 iterations. 
Models are characterized by their parameters (z M), layers (x), and head attentions (y), represented as (x, y, z).
Results reveal that increasing model size markedly boosts performance for all evaluated methods. 
Significantly, our HarmoDT demonstrates consistent superiority over MTDT and Prompt-DT across a range of model sizes.
Given that in HarmoDT and G-HarmoDT, the mask updating and gating module integrated into the decision transformer increase training costs and model size, we compare the size of trainable parameters and wall-clock time across different methods under the MT50 setting.
As shown in Table \ref{tab:parasizetime}, while G-HarmoDT requires slightly more parameters than MTDT, PromptDT, and HarmoDT due to the gating module, it has a much smaller model size than MGDT and MTDIFF. 
In terms of wall-clock training time, HarmoDT and G-HarmoDT take approximately 1.4\% and 4.6\% more time to update masks and the gating module compared to PromptDT, but significantly less time than MTDIFF.


\begin{figure*}[t!]
    \centering
    \subfigure[T-SNE visualization of optimal subspace in HarmoDT.]{
    \centering
    \label{fig:mask_tsne}
    \includegraphics[width=0.46\textwidth]{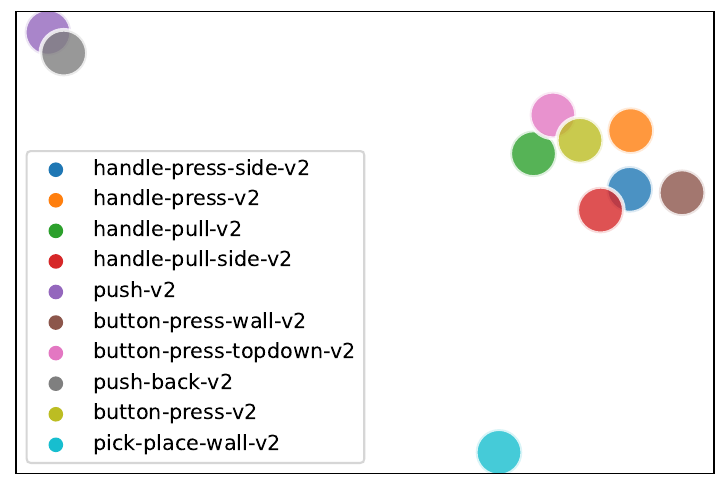}}
    \centering
    \subfigure[T-SNE visualization of grouping results in G-HarmoDT.]{
    \centering
    \label{fig:group_tsne}
    \includegraphics[width=0.405\textwidth]{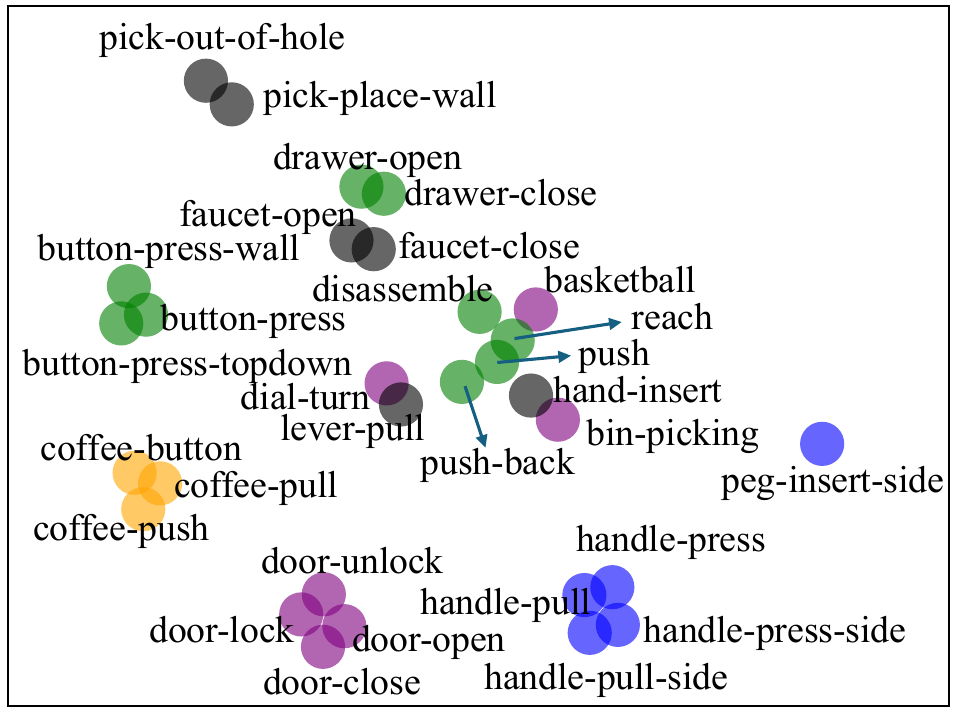}}
    \vspace{-0.2cm}
    \caption{T-SNE visualization of masks and grouping results on the 30 tasks of Meta-World benchmark under near optimal settings. The left panel shows the optimal subspace via masks learned by HarmoDT. The right panel illustrates grouping results of G-HarmoDT, where similar tasks are mapped to nearby positions in the semantic space using GloVe, and colors represent grouping results under 5Group. }
    \label{fig:ablation}
        \vspace{-0.3cm}
\end{figure*}

\subsubsection{Visualization of Mask and Grouping}
As shown in Figure~\ref{fig:mask_tsne}, we use t-SNE~\citep{tsne} to visualize the task masks generated by HarmoDT after training on 30 tasks from Meta-World benchmark.
Note that even small distances in the visualization can represent significant divergences in the original high-dimensional parameter space.
The visualization effectively showcases the relational dynamics of the task masks; closely related tasks such as `push-back-v2' and `push-v2' are positioned in proximity, while disparate tasks like `push-v2' and `pick-place-wall-v2' are distinctly separated.
This spatial arrangement underscores the efficacy of HarmoDT in delineating a harmony subspace tailored for each task.
As illustrated in Figure~\ref{fig:group_tsne}, we show the grouping results of G-HarmoDT.
Specifically, we project task names into the semantic feature space using GloVe~\citep{word2vec} and visualize it with t-SNE. 
The closer the distance, the more similar the tasks, and the colors correspond to our grouping results.
It can be observed that most tasks with close distances (having similar semantics) are assigned the same color, such as the "door" group, "handle" group, and "coffee" group. 
This demonstrates the effectiveness of our grouping mechanism, and KNN~\citep{knn1,knn2} is sufficient to achieve good grouping results.

\section{Conclusion}
In this study, we introduce the Harmony Multi-Task Decision Transformer (HarmoDT), a novel approach designed to discern an optimal parameter subspace for each task, leveraging parameter sharing to harness task similarities while concurrently addressing the adverse impacts of conflicting gradients.
By employing a bi-level optimization and a meta-learning framework, HarmoDT not only excels as a comprehensive policy in multi-task environments but also exhibits robust generalization capabilities to unseen tasks. 
To eliminate the need for task identifiers, we further design a group-wise variant (G-HarmoDT) that clusters tasks into coherent groups based on gradient information, and utilizes a gating network to determine task identifiers during inference.
Our rigorous empirical evaluations across a diverse array of benchmarks underscore our approach's superior performance compared to existing baselines, establishing its state-of-the-art effectiveness in MTRL scenarios.

\bibliography{main}
\bibliographystyle{plainnat}

\newpage
\onecolumn
\appendices

\section{Detailed Environment}
\label{sec:detailedenv}
\subsection{Meta-World}
The Meta-World benchmark, introduced by \citet{yu2020meta}, encompasses a diverse array of 50 distinct manipulation tasks, unified by shared dynamics. These tasks involve a Sawyer robot engaging with a variety of objects, each distinguished by unique shapes, joints, and connective properties. The complexity of this benchmark lies in the heterogeneity of the state spaces and reward functions across tasks, as the robot is required to manipulate different objects towards varying objectives. The robot operates with a 4-dimensional fine-grained action input at each timestep, which controls the 3D positional movements of its end effector and modulates the gripper's openness.
In its original configuration, the Meta-World environment is set with fixed goals, a format that somewhat limits the scope and realism of robotic learning applications. To address this and align with recent advancements in the field, as noted in works by \citet{sun2022paco, yang2020multi}, we have modified all tasks to incorporate a random-goal setting, henceforth referred to as MT50-rand. The primary metric for evaluating performance in this enhanced setup is the average success rate across all tasks, providing a comprehensive measure of the robotic system's adaptability and proficiency in varied task environments.

For the creation of the offline dataset, we follow the work by \citet{he2023diffusion} and employ the Soft Actor-Critic (SAC) algorithm \citep{sac} to train distinct policies for each task until they reach a state of convergence. Subsequently, we compile a dataset comprising 1 million transitions per task, extracted from the SAC replay buffer. These transitions represent samples observed throughout the training period, up until the point where each policy's performance stabilized. Within this benchmark, we have curated two distinct dataset compositions:
\begin{itemize}[leftmargin=*]
    \item \textbf{Near-optimal}: A dataset comprising 100 million transitions, capturing experience from random to expert-level performance (convergence) within SAC-Replay.
    \item \textbf{Sub-optimal}: A dataset comprising the initial 50\% of trajectories (50 million transitions) from the near-optimal dataset for each task, with a substantially reduced proportion of expert-level data.
\end{itemize}

\subsection{Unseen Tasks}
In our evaluation, we apply our approach to a diverse array of meta-RL control tasks, each offering distinct challenges to assess the performance and generalization capabilities of our model. The tasks are detailed as follows:
\begin{itemize}[leftmargin=*]
    \item \textbf{Cheetah-dir}: This task involves two distinct directions: forward and backward. The objective is for the cheetah agent to achieve high velocity in the assigned direction. The evaluation encompasses both training and testing sets, covering these two directions comprehensively to gauge the agent’s performance effectively.
    \item \textbf{Cheetah-vel}: Here, the task defines 40 unique sub-tasks, each associated with a specific goal velocity, uniformly distributed between 0 and 3 m/s. The agent's performance is assessed based on the $l_2$ error relative to the target velocity, with a penalty for deviations. For testing, 5 of these tasks are selected, while the remaining 35 are used for training purposes.
    \item \textbf{Ant-dir}: This task comprises 50 different sub-tasks, each with a goal direction uniformly sampled in a two-dimensional plane. The agent, an 8-jointed ant, is incentivized to attain high velocity in the designated direction. Of these, 5 tasks are earmarked for testing, with the rest allocated for training.
\end{itemize}

By evaluating our approach on these diverse tasks, we can assess its performance and generalization capabilities across different control scenarios.
The generalization ability of our approach is rigorously tested by examining the distribution of tasks between the training and testing sets, as outlined in Table \ref{tab:set_meta}. 
This experimental setup, as described in Section \ref{sec:exp}, adheres to the divisions specified, ensuring consistency in evaluation and enabling a comprehensive assessment of our approach’s adaptability and effectiveness across diverse control tasks.

\begin{table}[!ht]
    \caption{Training and testing task indexes when testing the generalization ability in meta-RL tasks }
    \label{tab:set_meta}
    \centering
    \begin{tabular}{ll}
      \toprule
      \multicolumn{2}{c}{Cheetah-dir} \\
      \midrule
      Training set of size 2 & [0,1] \\
      Testing set of size 2 & [0.1]\\
      \midrule
      \multicolumn{2}{c}{Cheetah-vel} \\
      \midrule
      Training set of size 35 & [0-1,3-6,8-14,16-22,24-25,27-39]\\
      Testing set of size 5 & [2,7,15,23,26]\\
      \midrule
      \multicolumn{2}{c}{Ant-dir} \\
      \midrule
      Training set of size 45 &  [0-5,7-16,18-22,24-29,31-40,42-49]\\
      Testing set of size 5 &  [6,17,23,30,41]\\
      \bottomrule
    \end{tabular}
\end{table}

\section{Hyper-parameters}
\label{sec:hypar}
This section details the training regimen implemented in our study. During the training phase, tasks are randomly selected for model refinement. Each training iteration is meticulously configured with a batch size of 256 and utilizes the Adam optimizer with a learning rate of 3e-4. The total number of training steps is set at 1e6. For a fair comparison, we adhere to the setups and recommendations in MTDIFF \citep{he2023diffusion}, setting the mask sparsity at $\mathrm{S}=0.2$, the mask update interval $t_m=5e3$, the upper bound of mask changes at $\eta_{max}=100$, the lower bound at $\eta_{min}=0$, and the controller $\lambda=10$.
The detailed process for selecting hyperparameters for HarmoDT is provided in Section \ref{sec:further}.
All methods are implemented using PyTorch on an NVIDIA GeForce 4090.

Our policy is built on a Transformer-based model, utilizing the minGPT open-source code\footnote{\url{https://github.com/karpathy/minGPT}}. The specific model parameters and hyper-parameters used in our training process are outlined in Table \ref{tab:model_parameter}.

\begin{table}[h]
\renewcommand{\arraystretch}{1.3}
\centering
  \caption{The model parameters of HarmoDT.}
  \vspace{0.1cm}
  \label{tab:model_parameter}
  \begin{tabular}{ll}
    \hline
    Parameter & Value \\
    \hline
    Number of layers            & 6 \\
    Number of attention heads   & 8 \\
    Embedding dimension         & 256 \\
    Nonlinearity function       & ReLU \\
    Batch size                  & 256 \\
    Context length $K$          & 20 \\
    Dropout                     & 0.1 \\
    Learning rate $\eta$              & 3.0e-4 \\
    Maximum iterations $E$             & 1e6 \\
    Sparsity $\mathrm{S}$              & 0.2 \\
    Minimum of mask changing $\eta_{min}$  & 0 \\
    Maximum of mask changing $\eta_{max}$  & 100 \\
    Balance factor   $\lambda$             & 10 \\
    Mask changing interval   $t_m$        & 5000 \\
    Threshold in Equation \ref{eq:unseen} & 25\\
    Warm up iterations $t_w$ for G-HarmoDT & 1e5 \\
    Number of groups $N^G$ for G-HarmoDT & 5 or 10 \\
    Number of input to gating network $ng$ for G-HarmoDT & 5\\
    \hline
  \end{tabular}
\end{table}

\section{ERK initialization}\label{app:ERK}
This section elucidates the utilization of the Erdős-Rényi Kernel (ERK), as proposed by \citet{ERK}, for initializing the sparsity in each layer of the model.
ERK tailors sparsity distinctively for different layers. 
In convolutional layers, the proportion of active parameters is determined by $\frac{n_{l-1}+n_l+w_l+h_l}{n_{l-1} \times n_l \times w_l \times h_l}$, where $n_{l-1}, n_l, w_l$, and $h_l$ represent the number of input channels, output channels, and the kernel's width and height in the $l$-th layer, respectively.
For linear layers, the active parameter ratio is set to $\frac{n_{l-1}+n_l}{n_{l-1} \times n_l}$, with $n_{l-1}$ and $n_l$ indicating the number of neurons in the $(l-1)$-th and $l$-th layers. 
ERK ensures that layers with fewer parameters maintain a higher proportion of active parameters.

\section{Baselines}\label{app:baseline}

We compare our proposed HarmoDT and G-HarmoDT with the following baselines.
\begin{enumerate}[leftmargin=*, label=\textit{\roman*}.]
    \item \textbf{MTBC}. We extend Behavior cloning (BC) to multi-task offline policy learning via network scaling and a task-ID conditioned actor that is similar to MTIQL.
    \item \textbf{MTIQL}. We extend IQL \citep{IQL} with multi-head critic networks and a task-ID conditioned actor for multi-task policy learning. The TD-based baselines are used to demonstrate the effectiveness of conditional generative modeling for multi-task planning.
    \item \textbf{MTDIFF \citep{he2023diffusion}}. MTDIFF is a diffusion-based method that integrates Transformer architectures and prompt learning for generative planning and data synthesis in multi-task offline settings.
    \item \textbf{MTDT}. We extend the Decision Transformer architecture \citep{DT} to learn from multitask data. Specifically, we pool large amounts of data as input to the Transformer architecture while retaining the original loss function.
    \item \textbf{PromptDT \citep{PDT}}. Prompt-DT, built on the Decision Transformer architecture, aims to learn from multi-task data and generalize policies to unseen tasks. Prompt-DT generates actions based on trajectory prompts and reward-to-go values.
    \item \textbf{MGDT \citep{lee2022multi}}. MGDT aims to control action generation to consistently produce highly-rewarding behaviors, which is necessary because the training datasets contain a mix of expert and non-expert behaviors, making it challenging to consistently generate expert behaviors directly from the model that imitates the data.
    \item \textbf{MTDT + PCGrad \citep{yu2020gradient} \& CAGrad \citep{liu2021conflict}}. We augment MTDT with gradient surgery techniques to resolve gradient conflicts. PCGrad projects a task’s gradient onto the normal plane of any other task's gradient that conflicts with it. CAGrad, on the other hand, minimizes the average loss function while leveraging the worst local improvement of individual tasks to regularize the algorithm's trajectory.
\end{enumerate}

In addition to offline methods, our analysis also encompasses a comparison with several online methodologies to provide a comprehensive evaluation of our approach. These include:

\begin{enumerate}[leftmargin=*, label=\textit{\roman*}., start=8]
    \item \textbf{CARE} \citep{sodhani2021multi}. This method utilizes additional metadata alongside a combination of multiple encoders to enhance task representation, offering a nuanced approach to multi-task learning.
    \item \textbf{PaCO} \citep{sun2022paco}. PaCO employs a parameter compositional strategy that recombines task-specific parameters, fostering a more flexible and adaptive learning process.
    \item \textbf{Soft-M} \citep{yang2020multi}.This approach focuses on developing a routing network that orchestrates the soft combination of various modules, thereby facilitating more dynamic learning pathways.
    \item \textbf{D2R} \citep{he2023not}. D2R innovatively employs disparate routing paths, enabling the selection of varying numbers of modules tailored to the specific requirements of each task, thereby enhancing the model's adaptability and efficiency.
\end{enumerate}

\vfill

\end{document}